\documentclass[%
 aip,
rsi,%
 amsmath,amssymb,
 reprint,%
]{revtex4-1}

\usepackage{graphicx}
\usepackage{dcolumn}
\usepackage{bm}

\begin{document}

\title{Understanding and mitigating noise in trained deep neural networks}

\author{N. Semenova}
\email{nadya.i.semenova@gmail.com}
\affiliation{D\'{e}partement d'Optique P. M. Duffieux, Institut FEMTO-ST,  Universit\'e Bourgogne-Franche-Comt\'e CNRS UMR 6174, Besan\c{c}on, France.}%
\affiliation{Department of Physics, Saratov State University, 83 Astrakhanskaya str., 410012 Saratov, Russia.}%

\author{L. Larger}
\affiliation{D\'{e}partement d'Optique P. M. Duffieux, Institut FEMTO-ST,  Universit\'e Bourgogne-Franche-Comt\'e CNRS UMR 6174, Besan\c{c}on, France.}%

\author{D. Brunner}
\affiliation{D\'{e}partement d'Optique P. M. Duffieux, Institut FEMTO-ST,  Universit\'e Bourgogne-Franche-Comt\'e CNRS UMR 6174, Besan\c{c}on, France.}%

\date{\today}

\begin{abstract}

Deep neural networks unlocked a vast range of new applications by solving tasks of which many were previously deemed as reserved to higher human intelligence.
One of the developments enabling this success was a boost in computing power provided by special purpose hardware, such as graphic or tensor processing units.
However, these do not leverage fundamental features of neural networks like parallelism and analog state variables.
Instead, they emulate neural networks relying binary computing, which results in unsustainable energy consumption and comparatively low speed.
Fully parallel and analogue hardware promises to overcome these challenges, yet the impact of analogue neuron noise and its propagation, i.e. accumulation, threatens rendering such approaches inept.
Here, we analyse for the first time the propagation of noise in parallel deep neural networks comprising noisy nonlinear neurons in fully connected layers.
We study additive and multiplicative as well as correlated and uncorrelated noise and develop analytical methods that predict the noise level in any layer of symmetric deep neural networks or deep neural networks trained with back propagation.
We find that noise accumulation is generally bound, and adding additional network layers does not worsen the signal to noise ratio beyond this limit.
Most importantly, noise accumulation can be suppressed entirely when neuron activation functions have a slope smaller than unity.
We therefore developed the framework for noise in fully connected deep neural networks implemented in analog systems, and identify criteria allowing engineers to design noise-resilient novel neural network hardware.

\end{abstract}

\maketitle

\section{Introduction}

A fundamental aspect of neural networks (NNs) is the propagation of information along the weighted connections between the countless nonlinear elements (neurons), making parallelism an essential prerequisite for an efficient implementation. 
Current digital computing architectures are, however, mostly serial and store connection weights in a memory spatially separated from the NN's nonlinear neurons.
This so called von Neumann bottleneck induces substantial energy penalties, and maximal computing performance can only be achieved if a NN's highly distributed topology is faithfully reproduced by the computing hardware. 
Only then overheads due to excessive information transduction are avoided: each neuron corresponds to a nonlinear component or circuit, each connection to a direct physical link.

Beyond computing-architecture aspects, there are also remarkable differences arising when considering principles of information encoding.
On one hand we have programs and computers based on symbolic, i.e. Boolean logic, for which the corruption of information has a disastrous impact.
A computer's binary voltage signal is therefore corrupted around once every 10 days, which at 1 GHz modulation bandwidth corresponds to one in $10^{24}$ \cite{Boahen2017}.
Information theory and thermodynamics link this error probability to an energy overhead dictated by the fundamental laws of physics.
NNs, on the other hand, leverage emergence in ensembles of nonlinear neurons.
Signals are therefore 'averaged' across numerous analogue states, and the impact of reduced signal finesse can \emph{potentially} be mitigated.
Already today special purpose NN chips, i.e. the newest generation of tensor and graphic processing units, allow low resolution (2-6 bit) computing, and algorithms of limited bit resolution are considered \cite{Gupta2015}.
Reducing the energy consumption of current NN hardware is therefore crucial already today, and research activity along these lines \cite{Hasler2013} has lately exploded.

Unfortunately, reducing bit-resolution mostly offers a linearly proportional energy saving.
Crucially, analogue signals are viable at such low resolution allows boosting energy efficiency ranging between 3 to 6 orders of magnitude \cite{Boahen2017}.
Novel analogue components such as lasers \cite{Brunner2013a}, memristors \cite{Tuma2016} and spin-torque oscillators \cite{Torrejon2017} have been shown to serve as excellent analogue neurons. 
Simultaneously, parallel networks based on holography \cite{Psaltis1990}, diffraction \cite{Bueno2018,Lin2018}, meshes of integrated Mach-Zehnder interferometers \cite{Shen2016}, wavelength division multiplexing \cite{Tait2017} and 3D printed optical interconnects \cite{Moughames2020,Dinc2020,Moughames-2-2020} have demonstrated fully parallel networking.
Finally, the concept of in-memory computing targets encoding a NN's topology based on tunable analogue circuits in electronic \cite{Wang2018,Lin2020,Xia2019} and photonic systems \cite{Feldmann2021}.

Current estimations only consider the impact of additive noise on the level of a single neuron.
This falls short of describing the reality of noise propagation in deep NNs (DNNs) in various aspects.
Firstly, neurons and connections can exhibit different types of noise.
Connections implemented via phase change materials \cite{Feldmann2021} exhibit thermal and parametric noise, electronic DNNs based on Boolean connections are sensitive comparator noise \cite{Moon2019}, electronic multi-layer perceptrons are sensitive to temperature and voltage fluctuations \cite{Janke2020}.
A wide range of analogue noisy hardware is therefore to be considered \cite{Dolenko1993, Misra2010, Dibazar2006, Soriano2015}.
Secondly, noise in DNNs will be sensitive to the connection statistics of each particular topology.
For example, fully connected layers strongly average signals, while only local connections do so significantly less.
Motivated by potentially substantial benefits, the central importance of noise for future NN hardware solutions is moving into focus.
Noise in analogue components mostly is additive and/or multiplicative and can be correlated and/or uncorrelated with other elements of the circuit.
Similar generalizations are possible for DNN topologies, where statistical methods rather than a description of each connection provide powerful tools.

The growing interest in analog DNN hardware increasingly identifies noise as a common presence.
In \cite{Moon2019} the authors consider an analog processor based on binary connections and study the effect of comparator noise, which is additive.
In \cite{Janke2020}, the authors consider an analog circuit of a multilayer perceptron and investigate the effect of temperature and voltage variation on accuracy, which corresponds to correlated additive noise.
Current studies exclusively focus on noise suppression in particular hardware \cite{Dolenko1993,Misra2010,Dibazar2006,Soriano2015,Frye1991}, but fail to capture the general principles of noise in analogue DNNs.
The description and analysis of noise in high-dimensional nonlinear systems is an established field of research in nonlinear dynamics \cite{Gailey1997,Shiino2001,Ichiki2007,Nakao2007}.
However, these do only consider the effect of static, i.e. not trained networks.

Here, we combine both approaches for the first time in order to develop the general theory of noise propagation in DNN that comprise noisy nonlinear neurons and have been trained for specific tasks.
In our previous work we considered the impact of noise on feed-forward and recurrent NNs only comprising linear neurons and networks with uniform, i.e. untrained connections \cite{Semenova2019}.
Here, we analytically investigate the general aspects of noise in DNNs comprising nonlinear neurons.
Importantly, we trained our DNNs according to standard error back-propagation. 

We demonstrate excellent accuracy of our analytical framework for DNNs trained in classification and analogue function approximation, hence for the majority of tasks in the machine learning context.
We find that nonlinear neurons decorrelate noise otherwise correlated across neuron populations.
Previously we identified such correlated noise as the most relevant perturbation, and noteworthy nonlinearity can therefore be beneficial for a NNs signal to noise ratio (SNR).
Our most important finding however is that noise propagation between consecutive network layers can be stopped completely, i.e. the accumulation of noise can be avoided.
We derive the general conditions for such noise-freezing and show that these are easily met in DNN concepts as well as hardware.
Finally, even without such noise freezing, a DNN's SNR is generally bound and increasing its depth does not further reduce its SNR below a limit.


\section{Neural networks and noisy neurons}

Noise at the output of a DNN imposes an upper limit of its computing accuracy.
For example, a perfect DNN model solving the MNIST problem without any misclassification would be limited in its misclassification to the DNN processors output signal quality.
For a low signal to noise ratio (SNR) of 10, such a system would create a misclassification 10\% of the time.
A detailed understanding of noise propagation is therefore of fundamental importance for next generation DNN hardware in order to combine energy-efficiency with computing misclassification .

\subsection{Deep neural networks}

\begin{figure}[t]
	\center{\includegraphics[width=1\linewidth]{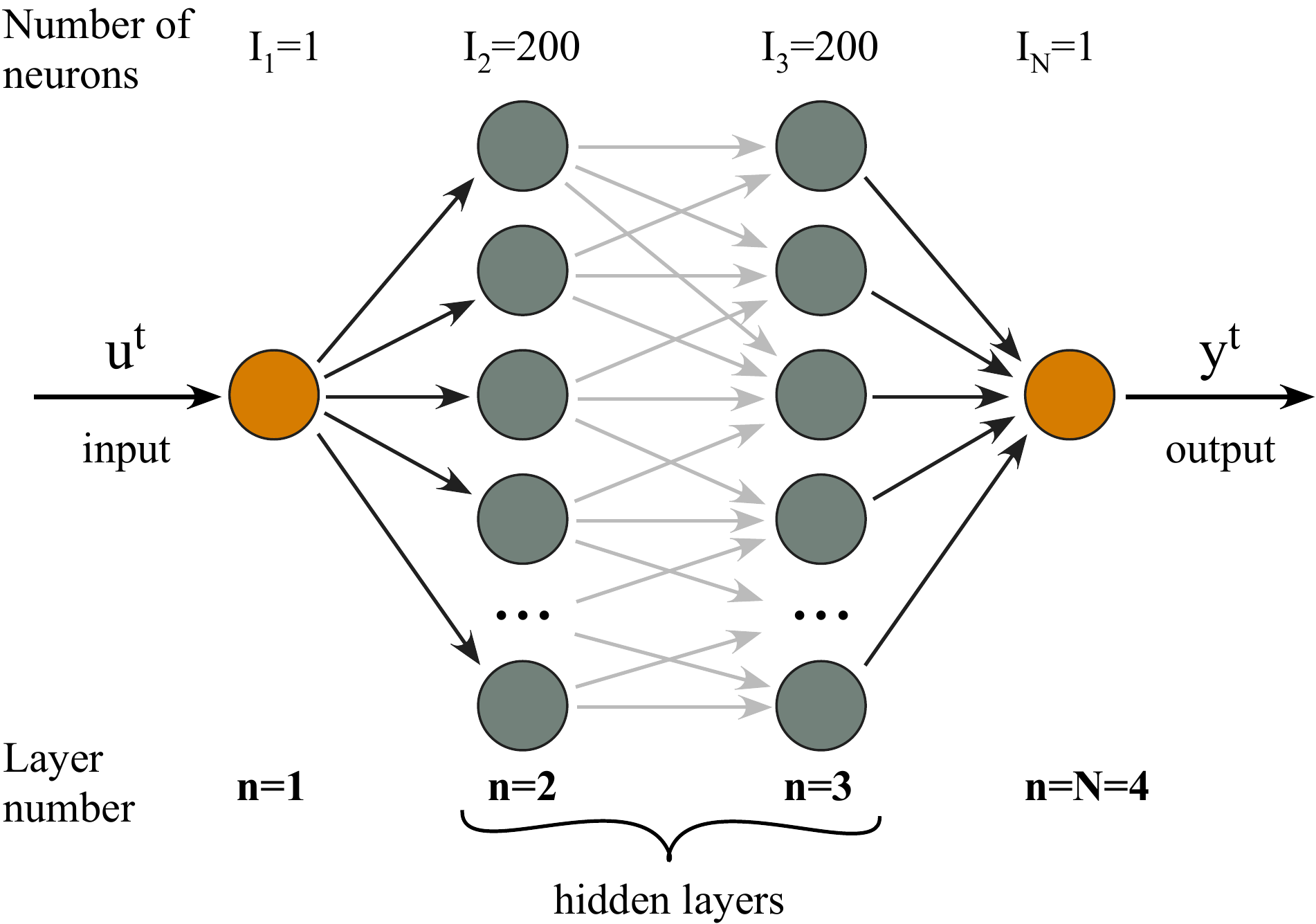}}
	\caption[]{Schematic illustration of a deep neural network.}
	\label{fig:SchemeDNN}
\end{figure}

DNNs may be found in multiple configurations, which can be categorized according to their connection topology. 
A simple DNN comprising input and output layers, each comprising a single linear neuron, plus two hidden layers hosting the nonlinear neurons is schematically illustrated in Fig.~\ref{fig:SchemeDNN}. 
Each neuron is identified by its layer index $n\in[1;N]$ and intralayer position $i\in[1;I_n]$ with $N$ and $I_n$ as the number of layers and neurons in a particular layer, respectively. 
At each integer instant $t\in[1;T]$ the network receives input signal $\mathbf{u}^t$.

In general, a neuron's internal state $\tilde{x}^t_{n,i}$ combines input from other neurons according to connectivity weight matrix $\mathbf{W}$ 
\begin{equation}\label{eq:NeuronFNN}
	\tilde{x}^t_{n,i} = \sum_{j=1}^{I_{n-1}} W^n_{i,j} y^t_{n-1,j} + b_{n,i},
\end{equation}
where $b_{n,i}$ is a constant bias and $y^t_{n-1,j}$ is the output signal of the $j$th neuron in the $(n-1)$th layer.
Equation (\ref{eq:NeuronFNN}) is valid for hidden and output layers, while for the input layer $\tilde{x}^t_{1,i}=u^t_{i}$. 
Internal states are typically transformed by a nonlinear function $f(\cdot)$, creating the neuron's output
\begin{equation}\label{eq:NodeOutFNN}
	x^t_{n,i} = f\left( \tilde{x}^t_{n,i} \right). 
\end{equation}
\noindent Here, we consider sigmoid $f(\tilde{x})=\frac{1}{2}+\frac{\alpha(\tilde{x}-0.5)}{2\sqrt{1+\alpha^2(\tilde{x}-0.5)^2}}$ as the neurons' nonlinearity.
We shifted the sigmoid to have its inflection point at $\tilde{x}=0.5$, and parameter $\alpha$ determines the slope.

A basic aspect of mapping with a nonlinear function $f(\cdot)$ is that a uniformly distributed input results in a non-uniformly distributed output. 
Nonlinear functions typically have sections with a low gradient, and for example the here used shifted sigmoid exhibits two horizontal asymptotes at $f(\tilde{x})=0$ and $f(\tilde{x})=1$.
It is in the vicinity of these asymptotes where the output signal is mainly concentrated, i.e. focused, and we refer to these zones as \textit{focusing points}.
For the rest of this manuscript we use a network with $N=4$, $I_1=I_4=1$ and $I_2=I_3=200$ as well as an input signal for length $T=1000$.

\subsection{Relevant types of noise}\label{sec:sect1_noise}

In experiments with analogue hardware NNs we carefully characterized noise present in the system \cite{Semenova2019, Andreoli2020}. 
In order to draw general conclusions we will consider additive and multiplicative noise.
Additive Gaussian noise is fundamentally present in any analogue circuit, for example due to the thermal noise of a resistor \cite{Boahen2017}.
When such a noisy resistor acts as reference in an operational amplifier circuit, its additive noise acts on the circuit's gain and hence is transformed into multiplicative noise.
Both are therefore perturbations commonly found in hardware \cite{Moon2019,Janke2020}.

\begin{figure}[t]
	\center{\includegraphics[width=1\linewidth]{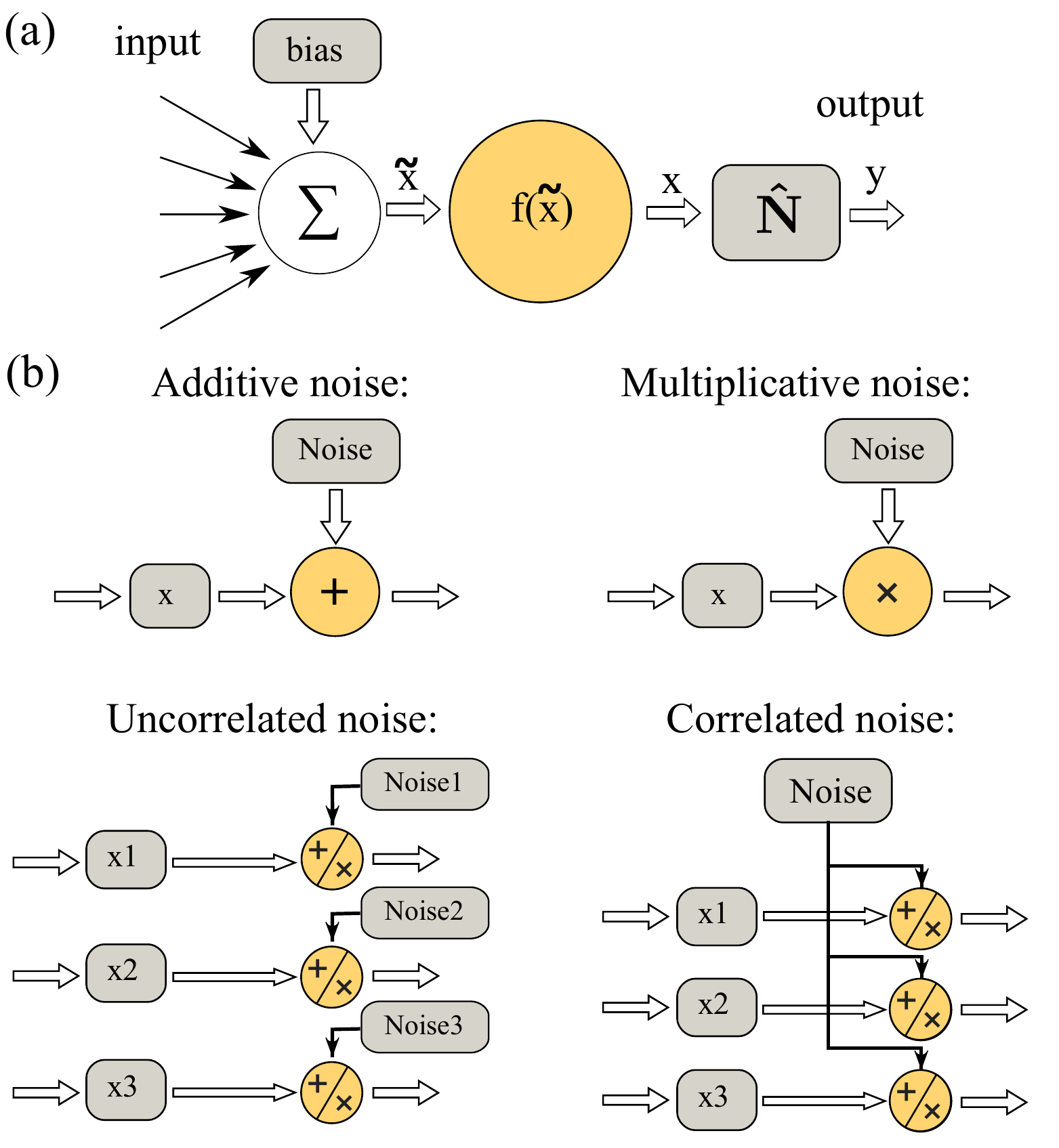}}
	\caption[]{(a) Schematic of a nonlinear neuron with noise. The nonlinear output $x=f(\tilde{x})$ is perturbed by noise operator $\hat{N}$ with additive and multiplicative noise. (b) Schematically illustration of different types of noise sources and their action upon a signal. From \cite{Semenova2019}.} 
	\label{fig:NoiseSources}
\end{figure}

Additive noise is added to noiseless signal $x_i$ according to $y_i=x_i+\sqrt{2D_A} \xi^A$, while multiplicative noise has a multiplicative relationship $y_i=x_i\cdot (1+\sqrt{2D_M} \xi^M)$. 
Here, $D_A$ and $D_M$ are the intensity for additive and multiplicative noise, respectively, while $\xi^{A|M}$ are the corresponding noise sources drawn from a normalized random Gaussian distribution.
Moreover, noise inside a group of neurons can be correlated or uncorrelated. 
In the latter case, noise affects each neuron differently at each instant $t$, while correlated noise identically perturbs the neurons within such a group at each $t$.
Combined, correlated additive $\xi^{C,A}$ and multiplicative $\xi^{C,M}$, uncorrelated additive $\xi^{U,A}$ and multiplicative $\xi^{U,M}$ noise perturb a signal according to
\begin{equation}\label{eq:one_neuron_noise}
	\begin{array}{c}
		y^t_{n,i}=\sqrt{2D^U_A}\xi^{U,A}_{n,i}+\sqrt{2D^C_A}\xi^{C,A}_{n}+ \\
		x^t_{n,i}\cdot\big( 1+\sqrt{2D^U_M}\xi^{U,M}_{ n,i}\big)\big( 1+ \sqrt{2D^C_M}\xi^{C,M}_{n}\big).
	\end{array}
\end{equation}
The canonical mechanisms of each of these is shown in Fig.~\ref{fig:NoiseSources}.
Here, we consider that noise is correlated for neurons contained within the same layer.

The main property of interest is the SNR encountered at a DNN's output. 
The SNR is obtained by dividing a signal's mean value by the square root of its variance. 
We numerically determine the variance by repeating the same input $k\in[1;K]$ times and hence this index is added to the neurons' noisy states $x^{t,k}_{n,i}$.
Here, we use $K=300$. The SNR for the $i$th neuron from the $n$th layer is
\begin{equation}\label{eq:SNR}
	\text{SNR}(y^t_{n,i}) = \frac{\text{E}(y^t_{n,i})}{\big( \text{Var}(y^t_{n,i})\big)^{1/2}},
\end{equation}
where $\text{E}(\cdot)$ and $\text{Var}(\cdot)$ are a variable's expectation value and variance, respectively.
For simplicity we define the SNR based on variance rather than standard deviation, which is the square root of variance, as this simplifies the following derivations and equations.

For a single neuron with Gaussian distributed and zero mean additive and multiplicative, correlated and uncorrelated noise (as in Eq.~(\ref{eq:NeuronFNN})), the expectation value and variance are
\begin{equation}\label{eq:EVar_one}
	\begin{array}{c}
		\text{E}(y^t_{n,i}) = \text{E}(x^t_{n,i}), \\
		\text{Var}(y^t_{n,i}) = \sigma^2_+ + \sigma^2_\times  \big(\text{E}(y^t_{n,i})\big)^2 + (1+\sigma^2_\times) \mathbf{\hat{F}}(\tilde{x}^t_{n,i}),
	\end{array}
\end{equation}
where $\sigma^2_+=2D^C_A+2D^U_A$, $\sigma^2_\times=2D^C_M+2D^U_M+4D^C_MD^U_M $ are the overall impact of additive and multiplicative noise, respectively. Operator $\mathbf{\hat{F}}$ corresponds to the variance of a random variable under the action of a nonlinear function $f(\cdot)$, in which case $\text{Var}(x^t_{n,i}) = \mathbf{\hat{F}}(\tilde{x}^t_{n,i})$, while for a linear neuron $\text{Var}(x^t_{n,i}) =\alpha^2 \text{Var}(\tilde{x}^t_{n,i})$. 

In the case of one isolated nonlinear neuron, see gray dashed data in Fig.~\ref{fig:SNR_DNN_Nonlinearities}, additive noise results in an SNR that increases linearly with the signal amplitude, and its slope is determined by noise intensity $D_A$. 
Purely multiplicative noise results in a constant level SNR for any output value. Mixed noise is characterised by a linear increase for small signal values, which for large signal amplitudes approaches the constant SNR-limit induced by multiplicative noise. The shape and law determining the SNR of individual neurons are therefore identical for linear and nonlinear neurons. Nonlinearity's only impact are the focusing points, and thus nonlinearity changes only the distribution of points along a single neuron's SNR curve. 
More details about individual noisy neurons in linear DNNs can be found in \cite{Semenova2019}, where we studied noise in feedforward and recurrent neural network with predefined, hence untrained coupling topologies. 
This simplification was made to separate the impact of noise and nonlinear function and we demonstrated that noise affects the final output signal differently depending on the type of noise. 

Variance $\text{Var}(\tilde{x}^t_{n,i})$ is the average noise impact from the previous layer $n-1$ and comprises contributions from correlated noise $N^C_n$, uncorrelated noise $N^U_n$ and noise $N^\mathrm{prev}_n$ from the layer preceding the previous one
\begin{equation}\label{eq:Var_3terms}
	\begin{array}{l}
		\text{Var}(\tilde{x}^t_{n,i}) = N^{C}_n+N^{U}_n+N^\mathrm{prev}_n\\
		
		N^{C}_n=I^2_{n-1}\mu^2(\mathbf{W}^n)\big( 2D^C_A+2D^C_M\mu^2(\text{E}(\mathbf{y}_{n-1})) \big)   \\
		
		N^{U}_n=I_{n-1}\eta(\mathbf{W}^n)\big( 2D^U_A+2D^U_M(1+2D^C_M)\eta(\text{E}(\mathbf{y}_{n-1})) \big)   \\
		
		N^\mathrm{prev}_n=I^2_{n-1}\mu^2(\mathbf{W}^n)\Big(1+2D^U_M\cdot\frac{\eta(\mathbf{W}^n)}{I_{n-1}\mu^2(\mathbf{W}^n)}\Big)\times \\
		\ \ \ \ \ \ \ \ \ \ \ \ (1+2D^C_M)\mu\big( \mathbf{\hat{F}}(\mathbf{\tilde{x}}^t_{n-1}) \big).
	\end{array}
\end{equation}
$N^\mathrm{prev}_n$ includes both, correlated and uncorrelated multiplicative noise, and operator $\mathbf{\hat{F}}$ as the influence of $f(\cdot)$.
Equations~(\ref{eq:Var_3terms}) include square of mean values $\mu^2(\cdot)$ and the mean of square  $\eta(\cdot)$ of coupling matrix $\mathbf{W}^n$, and depending on the particular type, noise propagation relies on $\mu^2(\cdot)$ and $I^2_{n-1}$ or $\eta(\cdot)$ and $I_{n-1}$ as well as the input signal's mean.
A DNN's topology therefore has a major influence on noise propagation due to its associated connection statistics.

\section{Noise in symmetric DNNs\label{sec:SymDNN}}

We initially focus on the influence of nonlinear function $f(\cdot)$, and hence start by considering a symmetric DNN where all connections $W^n_{ij}=1/I_{n-1}$ are uniform. 

\begin{figure}[t]
	\includegraphics[width=1\linewidth]{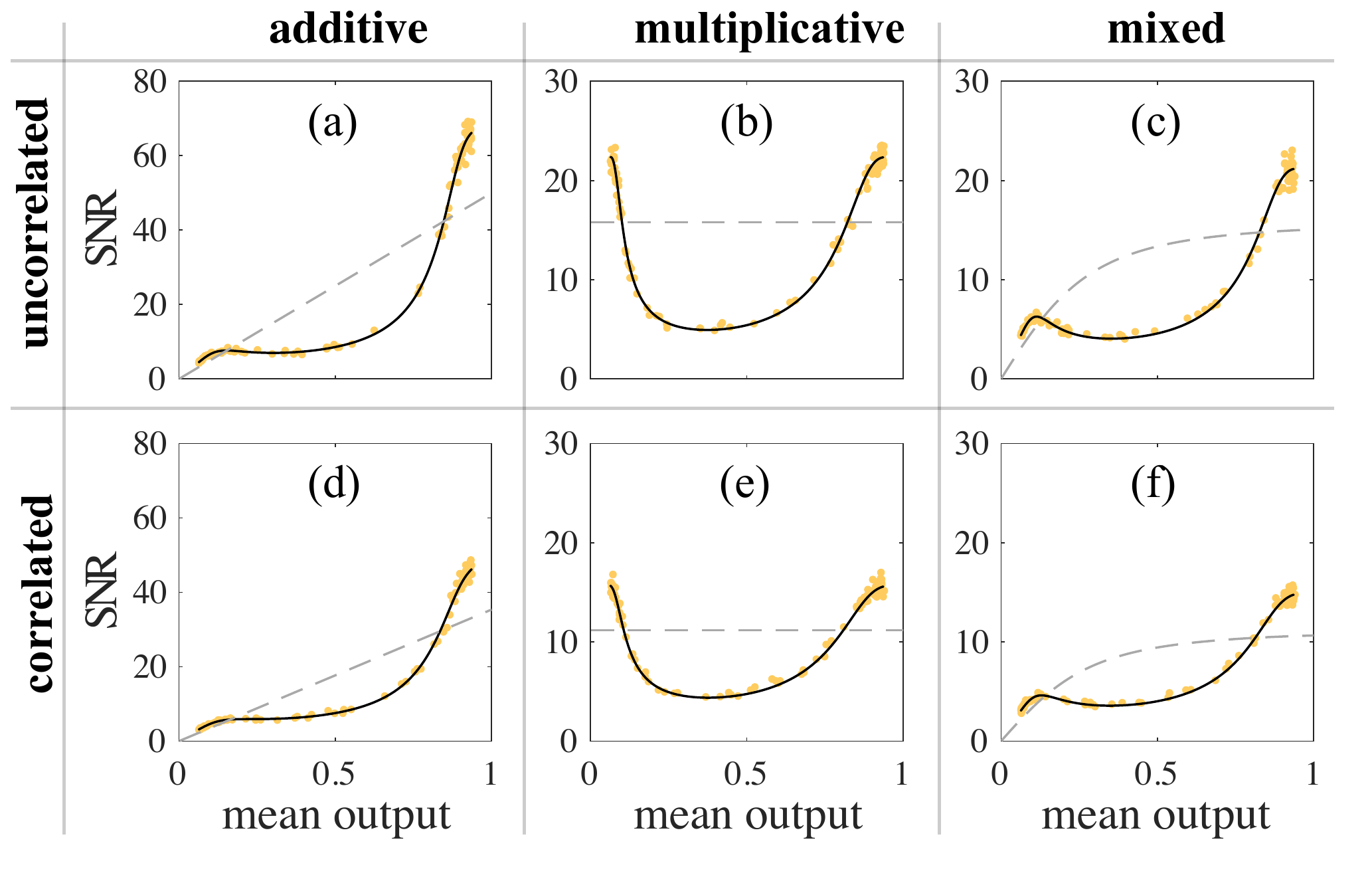}
	\caption[]{Signal-to-noise ratio (SNR) of the output in a nonlinear DNN (orange). The analytical prediction is given by black solid line. Gray dashed line show the SNR for a linear FNN with the same noise parameters, which is identical to the SNR of a single nonlinear neuron. Top panels shows the impact of uncorrelated noise, and bottom panels correspond to correlated noise. Here we used $D^U_A=D^C_A=10^{-4}$, $D^U_M=D^C_M=10^{-3}$ and $\alpha=4$.}
	\label{fig:SNR_DNN_Nonlinearities}
\end{figure}

As can be seen from Fig.~\ref{fig:SNR_DNN_Nonlinearities}(a--c), the SNR dependency for a DNN with our shifted sigmoid nonlinearity as activation function has two focusing points near $y=0$ and $y=1$. 
Orange data  correspond to numerically obtained SNR values for a nonlinear DNN for the case of additive (a), multiplicative (b) and mixed (c) uncorrelated noise, and the SNR significantly differs to a single nonlinear neuron. 
Correlated noise leads to a similar SNR dependency, and only slight quantitative differences occur if the correlated noise intensity is of a similar amplitude as the uncorrelated noise (Fig.~\ref{fig:SNR_DNN_Nonlinearities},(d--f)). 
Only disproportionately high noise intensities can lead to substantial qualitative modifications.

The SNR behaviour is determined by the general Eqs.~(\ref{eq:Var_3terms}), which can be simplified for symmetric DNNs.
If the number of neurons in the previous layer $I_{n-1}$ is large, the contribution of $N^C_n$ dominates over $N^U_n$ since $I^2_{n-1}\mu^2(\mathbf{W}^n)=1$ and $I_{n-1}\eta(\mathbf{W}^n)=1/I_{n-1}\ll1$.
Noise originating from the previous layer $N^{\mathrm{prev}}_{n}$ has terms with both multipliers, and for symmetric DNN term $I^2_{n-1}\mu^2(\mathbf{W}^n)$ dominates, hence $N^\mathrm{prev}_n\approx(1+2D^C_M)\mu(\mathbf{\hat{F}}(\mathbf{\tilde{x}}^t_{n-1}))$.
From this follows that $N^U_n$ is much smaller than both $N^C_n$ and $N^\mathrm{prev}_n$ in all hidden and the output layer
\begin{equation}\label{eq:Var_tilde_hidden}
	\text{Var}(\tilde{x}^t_{n,i}) \approx N^C_n+N^\mathrm{prev}_n, \ \ \ 2<n\le N.
\end{equation}
The input layer has no previous layer, and therefore
\begin{equation}
	\text{Var}(\tilde{x}^t_{n,i}) = N^C_n+N^U_n = \sigma^2_+ + \sigma^2_\times (\textbf{u}^t)^2, \ \ \ n=2.
\end{equation}

\subsection{First order approximation for nonlinear noise mixing}\label{sec:sec2_first}

A first order approximation of nonlinear noise-mixing operator $\mathbf{\hat{F}}$ depends on the first derivative of activation function $f(\cdot)$, the expectation value and variance of $\tilde{x}^t_{n,i}$ as $\mathbf{\hat{F}}(\tilde{x}^t_{n,i})\approx \big[ f'\big( \text{E}(\tilde{x}^t_{n,i}) \big)  \big]^2 \cdot \text{Var}(\tilde{x}^t_{n,i})$ (see Ref.~\cite{Benjamin2014}). 
If neurons are linear, as is the case for the input and output layers, this operator becomes $\mathbf{\hat{F}}(\tilde{x}^t_{n,i})=\alpha^2\text{Var}(\tilde{x}^t_{n,i})$. 
Otherwise, in a DNN's nonlinear hidden layers $1<n<N$, one obtains in first order
\begin{equation}\label{eq:F_simple}
	\begin{array}{c}
		\mathbf{\hat{F}}(\tilde{x}^t_{n,i})\approx \big[ f'\big( \text{E}(\tilde{x}^t_{n,i}) \big)  \big]^2 \cdot \text{Var}(\tilde{x}^t_{n,i}) \\
		\approx \Big[ f'\big( \text{E}(f^{-1}(y^t_{n,i})) \big)  \Big]^2 \cdot \text{Var}(\tilde{x}^t_{n,i}).
	\end{array}
\end{equation}
The expectation value of $\tilde{x}^t_{n,i}$ was substituted by $f^{-1}(y^t_{n,i})$ to maintain consistency with Eqs.~\ref{eq:Var_3terms}.

With the above simplifications Eqs.~(\ref{eq:SNR}--\ref{eq:Var_3terms}) were used to predict the impact of noise for symmetric DNNs and demonstrate an excellent agreement with numerical simulation, see  solid lines in Fig.~\ref{fig:SNR_DNN_Nonlinearities}. 
The final form of the set of equations for all layers with these simplifications is given in the Methods section. 

For symmetric DNNs with uniform matrices $\mathbf{\hat{F}}(\tilde{x}^t_{n,i})=\mathbf{\hat{F}}(\mathbf{\tilde{x}}^t_{n})$, $\text{Var}(\tilde{x}^t_{n,i})=\text{Var}(\mathbf{\tilde{x}}^t_{n})$ due to averaging; $\mu(\cdot)$ in Eq.~(\ref{eq:Var_3terms}) can therefore be neglected. According to Eqs. (\ref{eq:Var_3terms},\ref{eq:Var_tilde_hidden}) and (\ref{eq:F_simple}), the variance of $\mathbf{\tilde{x}}^t_{n}$ can be rewritten in a closed recurrent form based on the same variable from the previous layer:
\begin{equation}\label{eq:Var_tilde_rec}
	\begin{array}{cc}
		\text{Var}(\mathbf{\tilde{x}}^t_{n}) = N^C_n+ \\
		(1+2D^C_M)\Big[f'\big(E(f^{-1}(\mathbf{y}^t_{n-1}))\big)\Big]^2\text{Var}(\mathbf{\tilde{x}}^t_{n-1})
	\end{array}
\end{equation}
with initial state $\text{Var}(\mathbf{\tilde{x}}^t_{2}) = \sigma^2_++\sigma^2_\times(u^t)^2$. 
Here, we consider the realistic case where all noise intensities are $\ll$1, for which the contribution of $N^C_n$ has only a minor impact.
We can then neglect $N^C_n$, which allows us to focus on the consequence of network states propagating through a DNN's sequence of layers.
Leveraging the symmetric topology we redefine $\text{Var}(\mathbf{\tilde{x}}^t_{n})\approx S_n$ as a scalar sequence, and noise propagation of Eq.~(\ref{eq:Var_tilde_rec}) transforms to a geometric progression
\begin{equation}\label{eq:S_n}
	S_n = \Big[f'\big(E(f^{-1}(\mathbf{y}^t_{n-1}))\big)\Big]^2 \cdot S_{n-1}.
\end{equation}
If $f'(\cdot)\le 1$ on the entire interval of $y^t_n\in[0;1]$, then sequence $S_n$ decreases in general, otherwise it is increasing. 

\begin{figure*}[t]
	\centering{\includegraphics[width=1\linewidth]{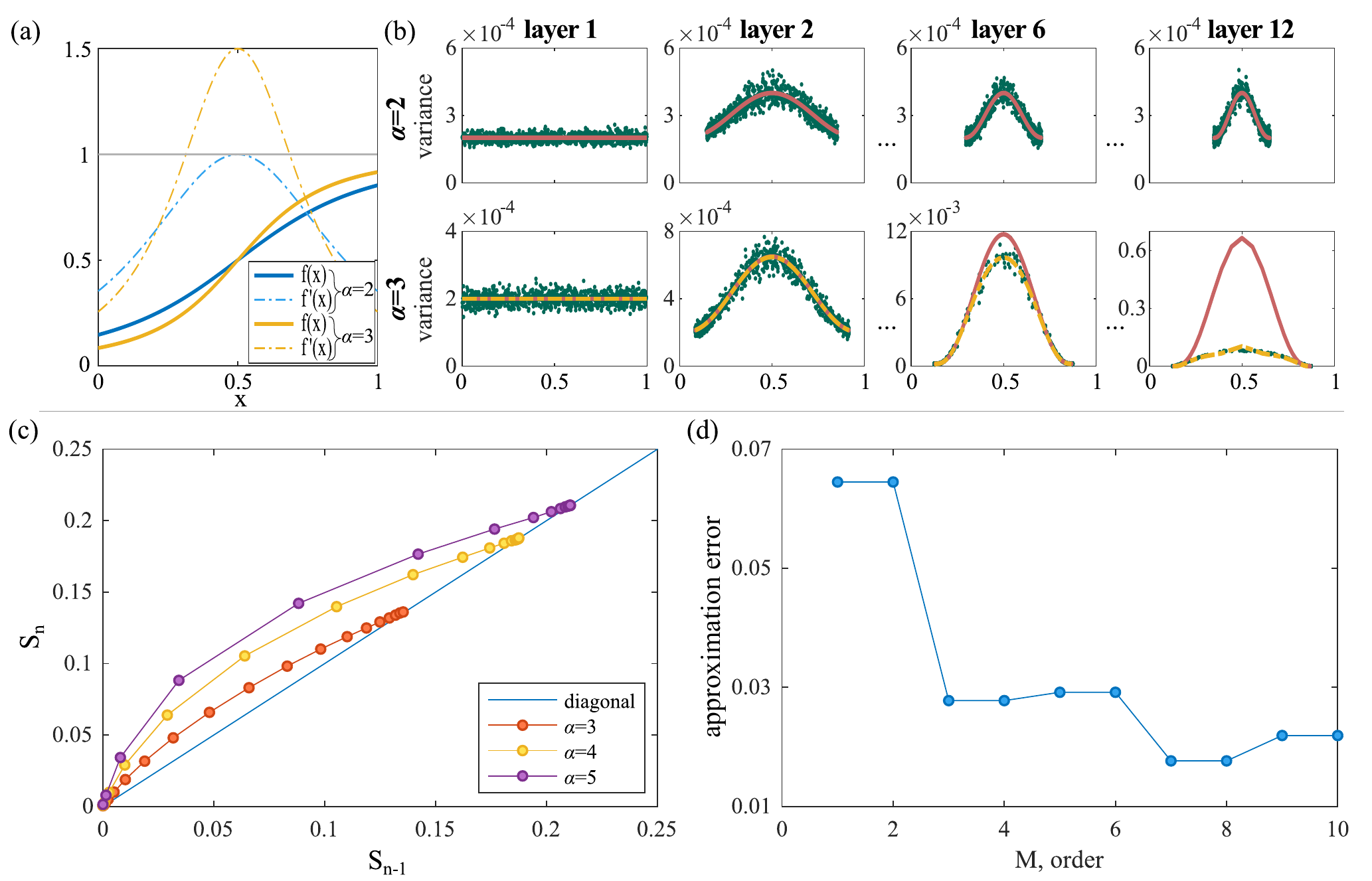}}
	\caption{(a) Activation functions and their first derivatives with $\alpha=2$ (blue lines) and $\alpha=3$ (orange lines). (b) Variance in different layers of a 12 layer DNN with $\alpha=2$ (top row) and $\alpha=3$ (bottom row). Red lines correspond to a first order approximation, yellow lines to a higher order approximation. (c) Lamerey diagram of $S_n$ for three $\alpha$ values is shown in panel. (d) Approximation error depending on the Taylor series's order $M$. The noise is uncorrelated additive $D^U_A=10^{-4}$.}
	\label{fig:large_alphas}
\end{figure*}

What this means in practice is that noise is kept from accumulating as long as the activation function's slope remains below unity.
The here used activation functions and their first derivatives are shown in Fig.~\ref{fig:large_alphas}(a) for $\alpha=2$ and 3, while Fig.~\ref{fig:large_alphas}(b) shows the variance inside a DNN until 12 layers for $\alpha=2$ (top row) and $\alpha=3$ (bottom row).
Red lines correspond to analytical predictions based on Eq.~(\ref{eq:F_simple}), and points to numerical simulations.
For our shifted sigmoid, the bifurcation in noise accumulation happens at $\alpha=2$; for $\alpha\le 2$ sequence $S_n$ decreases and noise propagation through the DNN's layers is prevented.
This is an extremely important insight.
Firstly, it shows that under not too stringent conditions a DNN's SNR does not worsen when increasing its depth; for $\alpha=2$ the variance remains essentially identical for all layers.
Secondly, nonlinearity and a networks depth are in a general competition when an analogue DNN's SNR is crucial. 

As can be seen from Fig.~\ref{fig:large_alphas}(b), using a more nonlinear activation function with $\alpha=3$ dramatically changes the situation: noise begins to accumulate and to strongly increase when adding hidden layers to the DNN. 
Furthermore, the now stronger nonlinearity results in a notable difference between the analytical description and numerical simulation, and this discrepancy increases as information propagates towards the deeper layers of the DNN. 
While the general behaviour of the variance curve is comparable to $\alpha=2$, the variance's amplitudes start to strongly diverge. 

We find that the speed at which variance $\text{Var}(\cdot)$ increases depends on the particular layer. 
Initially, growth is fast yet slows down for deeper layers and reaches a limit at some depth. 
This is well illustrated by a Lamerey diagram shown in Fig.~\ref{fig:large_alphas}(c), which demonstrates the numerically obtained evolution of $S_n$. 
As $S_n$ approaches the diagonal $S_n=S_{n-1}$ growth slows down and noise amplitudes from one layer to the next remain constant, i.e. stop increasing. 
We evaluated several $S_n$ with $\alpha>2$, which all reached some limit, each at different depth and corresponding $S_n$ limit. 
Two main rules emerged: the larger $\alpha$, the larger the final $S_n$, yet also the fewer layers are required to reach the particular limit.

Though included, the data of $S_n$ in Fig.~\ref{fig:large_alphas}(c) for $\alpha<2$ is hardly visible. 
In this case the scale of $S_n$ does not change with depth of the network, and $S_n$ remains essentially constant at the single neuron level of $10^{-3}$. 
All $S_n$ values are therefore located near the origin of Fig.~\ref{fig:large_alphas}(c) for all $\alpha\leq2$, and the SNR does not significantly surpass a single neuron's SNR for the generally valid assumption $I_n\gg 1$ for $n>1$. 

\subsection{Higher order approximation}\label{sec:sec2_high}

In order to explain the difference in Fig.~\ref{fig:large_alphas}(b) between our first order analytical approximation and numerical simulations, it is necessary to return to the reason why coefficient $f'\big(E(f^{-1}(\mathbf{y}^t_{n-1}))\big)$ appeared in the first place. 
It determines the variance of nonlinear function $f(\cdot)$ for random variables. 
For a random variable $X$ with mean value $\mu(X)$ and probability density distribution $p_X(x)$, $Y=f(X)$ has variance $\text{Var}(Y)=E(Y^2)-\big( E(Y) \big)^2$ with $E(Y)=\int^\infty_{-\infty}f(x)p_X(x)dx$. 
Function $f(\cdot)$ can in general be approximated using a Taylor series of $M$th order. 
Equation~(\ref{eq:F_simple}), however, included only the Taylor series's  1st order, i.e. until the first derivative. 
After several monotonous transformations, given in the Methods section, we arrive at the variance with $M$th order approximation
\begin{equation}\label{eq:F_M}
	\begin{array}{c}
		\mathbf{\hat{F}}(\mathbf{\tilde{x}}^t_{n})\Big|_M=\sum\limits^M_{m=1}T^2_m(V_{2m}-V^2_m)\\
		+2\sum\limits^M_{m=1}\sum\limits^{m-1}_{i=1}T_m T_i(V_{m+i}-V_mV_i),
	\end{array}
\end{equation}
or in a recurrent form based on the variance of the preceding order
\begin{equation}\label{eq:F_M_rec}
	\begin{array}{c}
		\mathbf{\hat{F}}(\mathbf{\tilde{x}}^t_{n})\Big|_M=\mathbf{\hat{F}}(\mathbf{\tilde{x}}^t_{n})\Big|_{M-1}    +     T^2_M\cdot(V_{2M}-V^2_M)\\
		+2T_M\sum\limits^{M-1}_{i=1}T_i(V_{M+i}-V_MV_i),
	\end{array}
\end{equation}
where $T_m=\frac{1}{m!}f^{(m)}(\mu)(x-\mu)^m$ is the $m$th order term of the Taylor series approximated nonlinearity contribution and $V_m=\int^\infty_{-\infty}(x-\mu)^m p_X(x)dx$ is the $m$th moment of the probability density distribution $p_X(x)$.

Due to $S_n\approx\mathbf{\hat{F}}(\mathbf{\tilde{x}}^t_{n})$, the higher order approximations of $S_n$ strongly depend on probability density $p_X$ and expectation value $\mu$. 
These are known for the first two orders: $V_1$ is equal to zero and $V_2=\text{Var}(\mathbf{\tilde{x}}^t_{n})$. 
The preceding terms need to be specifically calculated. 

Figure \ref{fig:large_alphas}(b) shows $S_n$ obtained via the approximation of $\mathbf{\hat{F}}$ until the 15th order as yellow lines. Figure~\ref{fig:large_alphas}(d) shows the discrepancy between our analytical approximation based on different orders and the full-nonlinearity based numerical solutions at the point of the largest variance, found at output $\tilde{x}=0.5$, which is also where the largest discrepancy between analytic and numerical solutions are found.
Interestingly, differences between analytical approximation and numerical calculations are arranged in pairs: every second point provides the same accuracy as an approximation based on its preceding order. 
This is because every even derivative of $f(\cdot)$ is zero at the point of inflection at $\tilde{x}=0.5$, and the higher order approximation following an uneven order $M$ therefore has the same value.

In general, for $M\to \infty$ a Taylor series approximates a function on the whole set of values, not only in a small neighbourhood.
However, in our setting only interval $[0;1]$ is of relevance, and for some $M$ a better global approximation simultaneously approximates with less accuracy inside $[0,1]$.
For this reason, Fig.~\ref{fig:large_alphas}(d) shows a globally decreasing sequence, but some higher orders might result in a worse approximation than their preceding orders.


\section{Noise in fully trained DNNs\label{sec:TrainedFNN}}

Training DNNs is based on changing connection matrices and biases, and naturally weight matrices and their statistical
properties differ depending on tasks and training methods. 
Crucially, due to the typically large number of neurons in hidden layers, weight as well as bias distribution statistics after training are sufficient to describe noise propagation in trained nonlinear DNNs. 

We therefore consider a coupling matrix as a set of random variables, whose statistical properties such as mean square and squared mean are essential for variance calculations.  
The analysis for uniform networks given in Sect.~\ref{sec:SymDNN} remains mostly valid for trained networks, except for an admittedly cumbersome modification to $S_n$
\begin{equation}\label{eq:Sn_trained}
	\begin{array}{l}
		S_n\approx I^2_{n-1}\mu^2(\textbf{W}^n)\cdot[2D^C_A+2D^C_M\mu^2\{\text{E}(\mathbf{y}^t_{n-1})\}]+\\
		I_{n-1}\eta(\textbf{W}^n)\cdot[2D^U_A+2D^U_M(1+2D^C_M)\cdot\eta(\text{E}(\mathbf{y}^t_{n-1}))]  \\
		+I_{n-1}\eta(\textbf{W}^n)(1+\sigma^2_\times) \cdot\eta\{f'(f^{-1}(\text{E}(\mathbf{y}^t_{n-1})))\}S_{n-1},
	\end{array}
\end{equation}
with initial value $S_2=\sigma^2_++\sigma^2_\times\mu^2(\mathbf{u}^t)$ as for symmetric DNNs.
After training a DNN model, the squared mean $\mu^2(\textbf{W}^n)$ and mean square $\eta(\textbf{W}^n)$ of connection matrices remain static and are known, yet $\text{E}(\mathbf{y}_{n-1})$ and $f'(f^{-1}(\text{E}(\mathbf{y}_{n-1})))$ needs to be approximated as it depends on the particular input information. 
The relevant SNR-contributions related to the expectation values of neurons then are
\begin{equation}\label{eq:mu_eta_integrals}
	\begin{array}{cc}
		\mu(\text{E}(\mathbf{y}_n)) = \int\limits^\infty_{-\infty} f(z)p^*_n(z)dz \\
		\eta(\text{E}(\mathbf{y}_n)) = \int\limits^\infty_{-\infty} \big(f(z)\big)^2p^*_n(z)dz \\
		\eta(f'(f^{-1}(\text{E}(\mathbf{y}_n)))) = \int\limits^\infty_{-\infty} \big(f'(z)\big)^2p^*_n(z)dz,
	\end{array}
\end{equation}
where $p^*_n(z)$ is the probability density of neuron states $\tilde{\mathbf{x}}_n$ under the influence of the activation function $f(\cdot)$. 
When investigating different computational tasks we found that distribution $p^*_n(\cdot)$ can be well approximated only utilizing the weights' statistics after training and the input data. 
Details are given in Methods \ref{sec:Methods_pdf}.

In order to demonstrate the validity of our approach approach works for DNNs regardless of the specific task, we trained three noiseless networks for different tasks: DNN A - classification of handwritten digits (MNIST digits); DNN B - classifying clothing images (MNIST fashion); DNN C - analogue target approximation while predicting a chaotic Mackey-Glass sequence \cite{Jaeger2004}. 
Furthermore, DNN A is trained and tested using a constant, hence not optimized bias for each neuron, while network B has optimized biases in each layer. 
The resulting statistics of biases and coupling matrices are given in Appendix A.
Connection weights and biases are optimized during training based on an Adam optimization algorithm, and their values can be positive and negative.
All three networks were trained using the Keras neural-network library \cite{Keras}.
To build our network we have used brain.js \cite{Brainjs}. 

Deep neural networks have demonstrated excellent performance for image classification \cite{LeCunSite}. 
To enable comparison with the previous sections, we continue using two hidden layers with 200 neurons. 
Crucially, the analytical approximation can readily be extended to other topologies, i.e. deeper networks with larger layers or DNNs with a convolutional topology. 
Here we use a sigmoid  without shift $f(\tilde{x})=1/(1+e^{-\tilde{x}})$ as activation function. 
For training and testing handwritten digit recognition we used the MNIST database containing more than 10000 images. 
Each image consists of 28$\times$28 pixels in a gray scale, resulting in 784 input nodes. 
For classifying digits 0--9 the system has 10 output neurons, and the result is determined by the output neuron with the largest value. 
The trained DNN A has an averaged training error of $\approx 1\%$ and testing error of $\approx 1\%$ averaged across all digits. 

\begin{figure}[h]
	\center{\includegraphics[width=1\linewidth]{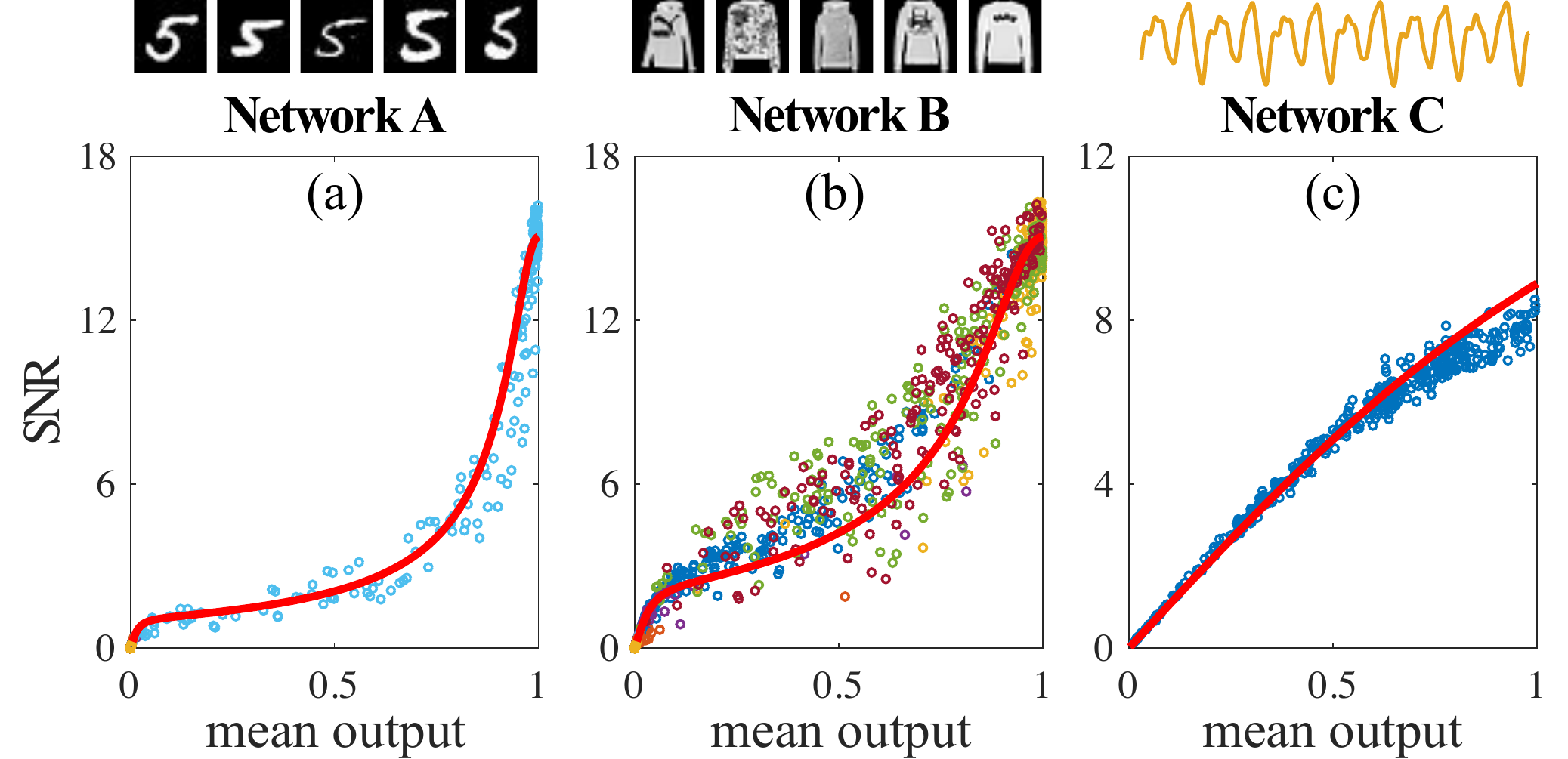}}
	\caption[]{SNR for the output layers of three networks: A -- digits recognition (a) tested on 1000 images of digit ``5''; B -- clothing detection (b) tested on 1000 images of ``pullowers''; C -- prediction of Mackey--Glass system realization . The solid lines correspond to analytical prediction. The analytic approximation depends on approximation $p^*(z)$}
	\label{fig:trained_nets}
\end{figure}

The SNR dependency is numerically obtained as before using the trained network topologies, yet now under noisy conditions. 
Each input image is repeated $K$ times to calculate the expected value and the standard deviation. 
Figure \ref{fig:trained_nets}(a) shows the SNR in the final layer for different example images of digit ``5'' as input, which is comparable to the SNR for all 10 different digits. 
Blue data in Fig.~\ref{fig:trained_nets}(a) corresponds to the class of digit ``5'', and data is distributed across the entire range.
Yellow data data corresponds to the remaining 9 other classes, and a strong focus to small mean amplitudes illustrates successful classification.
The red solid line is our analytical prediction based on a first order approximation, which is in very good agreement with numerical simulations.

The similar network but now also optimizing biases during training is considered for the MNIST-fashion data set. 
This data set contains images of clothes and shoes with T-shirt/top, Trouser,  Pullover, Dress, Coat, Sandal, Shirt, Sneaker, Bag, Ankle boot as labels.
Again, input data is a 28x28 pixel image. 
Figure ~\ref{fig:trained_nets}(b) shows the output layer SNR for 1000 images of pullovers. 
The analytically predicted dependency (solid line) works well in this case too.
In this task DNN B obtained a $5\%$ training and $10\%$ testing error rate,

Feed forward networks usually are used for classification tasks. 
In order to emphasize the generality of our analytical framework, we consider the following example for analogue signal prediction. 
The DNN's input nodes receive the 100 previous points of a chaotic Mackey--Glass system\cite{MackeyGlass1997} at each time $t$. 
The network's output is then tasked to predict the next point of the chaotic sequence. 
The network therefore has 100 input neurons and one output neuron, and the hidden layers contain 200 neurons, as previously.
After training using Keras the network achieves 87\% prediction accuracy. 
Figure~\ref{fig:trained_nets}(c) shows the SNR in the output layer, and the red solid lines the analytically obtained SNR dependency, which is in good agreement with our numerical simulation.

\section{Discussion and Conclusion\label{sec:DiscussionConclusion}}

In this article we studied the effect of internal noise on the output of DNNs.
Several models of different complexity were considered. 
The general features depending on the DNN's nonlinear activation function and depth were shown for the symmetric DNNs with full and uniform connectivity. 
The SNR depends on both the nonlinear function and the type of noise involved.
The analytical methods approximate the noise level in all layers very well. 

All obtained results were validated for three trained DNNs used for digits recognition, clothing detection and time series prediction, and an excellent agreement of the obtained results with the prediction was obtained regardless of DNN topology. 
Crucially, our analytical approximations only depend on connection and biase statistics. 
Our method therefore is agnostic to the detailed, local configuration of neurons and connections. 
This is crucial when it comes to approximating the SNR performance of a real, hardware implemented DNN. 
One can employ a numerical toy-model with the general parameters of the hardware system such as activation functions and their noise amplitudes and characteristics, which usually are readily available. 
The toy model is then trained, and while the local configuration of a hardware DNN will not correspond to the trained toy model \cite{Andreoli2020,Freiberger2020}, bias and connection statistics are likely to be similar.

Finally, the proposed analytical methods to trained networks and the good agreement between theory and numerical simulations demonstrated in  (Fig.~\ref{fig:trained_nets}) shows the robustness of our methods.
Our approach does not depend on or requires knowledge of details of training procedure, but instead only relies on the statistics of connection matrices after the training procedure.
Based on these, the statistical distribution of the state vector in each DNN layer can be approximated, see Section \ref{sec:Methods_pdf} in the appendix for more information.
Our analytical approach should therefore be widely applicable for DNNs where such approximations hold.
Very small DNNs or DNNs with very sparse connectivity could evade such an abstraction based on statistical methods.
One future direction is therefore a detailed evaluation of such mostly locally connected models. 

A very interesting and highly relevant result is the SNR's behaviour when the network depth is increased. 
We initially (and quite intuitively) expected to find the noise level to continue increasing as a function of the number of layers. 
However, we demonstrated that the output layer's SNR is generally bound by a lower limit. 
A DNN's SNR does therefore not worsen when layers are added once this limit is reached, and the exact value is determined mainly by the activation function's degree of nonlinearity. 
Most importantly, we find (i) that nonlinearity and weight statistics efficiently avoid accumulation of correlated noise, and (ii) that propagation and accumulation of noise from one to the next layer is fully suppressed when using activation functions with a slope smaller unity. 
These are very easy and realistically satisfiable requirements, and the output SNR for such hardware DNNs is essentially identical to the one of a single neuron. 
It therefore should be plausible and realistic to design and implement analog hardware DNNs which are very robust to internal noise.
Finally, our model also elegantly allows including noise in the input data simply by assigning its properties to the first layer neurons, which in normal concepts are linear and noiseless.

Future work needs to consider other DNN topologies and to elaborate if our general and promising findings can be confirmed for potentially more restricted connection matrices. 
Further efforts should focus on deriving the accumulative SNR for a particular task, hence to obtain to obtain an upper accuracy limit imposed by the DNN's internal noise. 
Finally, our work provides important insight towards DNN topologies with enhanced noise resilience.
The identification of architectures that provide noise robustness and concepts that leverage potential benefits of noise on learning learning \cite{Murray1991} is an important future line of research.

Finally, our work raises questions regarding the role of nonlinearity and hierarchy in biological NNs or more neuro-inspired hardware \cite{Neckar2019,Yang2021,Yang2021b}.
Original biological NNs are considered very noisy.
It is an interesting and open question if noise accumulation is only prevented by local neural functionality, as for example the threshold of integrated and fire neurons or simply their nonlinear transfer function.
An alternative approach could leverage learning to inherently optimize the network's topology to a regime that increases noise-robustness.
Noise and robustness related learning are important current challenges in next generation hardware. 
Efficient concepts to solve the credit assignment problem in noisy systems, such as dendritic event-based processing \cite{Yang2021b}, have been demonstrated.
Another challenge is mitigating the fault of individual connections or neurons, and the concept of fault-tolerant spike routing in spiking neural networks provides a robust and scalable concept for context dependent learning \cite{Yang2021c}.
In general, our work combined with the recent results on learning concepts \cite{Yang2021b,Yang2021c} opens further interesting avenues to scientifically explorer for leveraging local learning rules.

\section{Acknowledgments\label{sec:Acknowledgments}}

The authors acknowledge the support of the Region Bourgogne--Franche-Comté.
This work was supported by the EUR EIPHI program (Contract No. ANR-17-EURE-0002) and the Volkswagen Foundation (NeuroQNet II).
N.S. is supported by Russian Science Foundation (Project No. 21-72-00002).

\section{Methods\label{sec:Methods}}

\subsection{Noise propagation in symmetric FNNs}\label{sec:Methods_analytics_DNN}

The output of a single neuron is described by Eq.~(\ref{eq:one_neuron_noise}). The expectation value of the output from the $i$th neuron of the $n$th layer is
\begin{equation}
	\text{E}[y^t_{n,i}] = \text{E}[x^t_{n,i}].
\end{equation}
According to general rules of mathematical operations for uncorrelated random variables $X$ and $Y$, one obtains $\text{Var}[X+Y]=\text{Var}[X]+\text{Var}[Y]$ and $\text{Var}[X\cdot Y]=\text{Var}[X]\cdot\big(\text{E}[Y]\big)^2+\Big( \big(\text{E}[X]\big)^2+\text{Var}[X] \Big)\cdot\text{Var}[Y]$. 
When $X$ is multiplied on a constant $C$, the variance becomes $\text{Var}[C\cdot X]=C^2\text{Var}[X]$. 
All considered noise sources $\xi$ have zero expected value and a variance equal to $1$. 
The final variance of the noise source is controlled by the corresponding noise intensity, and \\$\text{Var}(\sqrt{2D^U_A}\xi^{U,A}_{n,i})=2D^U_A$. 
Taking the above into account, the variance of $y^t_{n,i}$ can be calculated as:
\begin{equation}\label{eq:Var_y}
	\begin{array}{c}
		\text{Var}[y^t_{n,i}]=2D^U_A+2D^C_A+\\
		\text{Var}\big[ x^t_{n,i}(1+\sqrt{2D^C_M}\xi^{C,M}_{n})(1+\sqrt{2D^U_M}\xi^{U,M}_{n,i}) \big] \\
		=2D^U_A+2D^C_A+2D^U_M\big(\text{E}[x^t_{n,i}]\big)^2+ \\
		(1+2D^U_M) \text{Var}\big[ x^t_{n,i}(1+\sqrt{2D^C_M}\xi^{C,M}_{n,i}) \big] \\
		=  2D^U_A+2D^C_A+
		(2D^U_M+2D^C_M+4D^U_M D^C_M)\times \\
		\big(\text{E}[x^t_{n,i}]\big)^2+
		(1+2D^U_M)(1+2D^C_M)\text{Var}[x^t_{n,i}].
	\end{array}
\end{equation}
According to $\sigma^2_+$ and $\sigma^2_\times$ introduced in Sect.~\ref{sec:sect1_noise}, Eq.~(\ref{eq:NeuronFNN}) can be simply rewritten in the form of Eq.~(\ref{eq:EVar_one}).

The variance of the combined signal, propagating from layer $n$ to layer $(n+1)$, $\tilde{x}^t_{n+1,i}$ is
\begin{equation}\label{Var_x_tilde}
	\begin{array}{c}
		\text{Var}[\tilde{x}^t_{n+1,i}] = \text{Var}\Big[ \sum\limits^{I_n}_{j=1}W^{n+1}_{ij}y^t_{n,j} + b_{n+1,i}  \Big] = \\ \text{Var}\Big[ \sum\limits^{I_n}_{j=1}W^{n+1}_{ij}y^t_{n,j}\Big] .
	\end{array}
\end{equation}
Some terms of $y^t_{n,j}$ are independent of index $j$, and they therefore can be moved outside the sum. 
Moreover, almost all variables in Eq.~(\ref{Var_x_tilde}) depend on time. 
We can therefore neglect index $t$ and include it only in the final equations.
\begin{equation}\label{Var_x_tilde2}
	\begin{array}{c}
		\text{Var}[\tilde{x}_{n+1,i}] = \text{Var}\Big[  (1+\sqrt{2D^C_M}\xi^{C,M}_n)\times \\
		\sum\limits^{I_n}_{j=1}W^{n+1}_{ij}
		(1+\sqrt{2D^U_M}\xi^{U,M}_{n,j})x_{n,j} + \\
		\sqrt{2D^U_A}\sum\limits^{I_n}_{j=1}W^{n+1}_{ij}\xi^{U,A}_{nj} + 
		\sqrt{2D^C_A}\xi^{C,A}_{n}\sum\limits^{I_n}_{j=1}W^{n+1}_{ij} \Big]\\
		= 2D^U_A\sum\limits^{I_n}_{j=1}\big(W^{n+1}_{ij}\big)^2+
		2D^C_A\Big(\sum\limits^{I_n}_{j=1}W^{n+1}_{ij}\Big)^2 + \\
		2D^C_M\Big(\text{E}\big[\sum\limits^{I_n}_{j=1}W^{n+1}_{ij}x_{n,j}\big]\Big)^2 + (1+2D^C_M)\times \\
		\text{Var}\Big[  \sum\limits^{I_n}_{j=1}W^{n+1}_{ij}(1+
		\sqrt{2D^U_M}\xi^{U,M}_{nj})x_{n,j}  \Big]
	\end{array}
\end{equation}

The connection matrix $\mathbf{W}^{n+1}$ has the mean of square value $\eta(\mathbf{W}^{n+1})=\frac{1}{I_n I_{n+1}}\sum^{I_{n+1}}_{i=1} \sum^{I_{n}}_{j=1} \big( W^{n+1}_{ij}\big)^2$ and mean value \\ $\mu(\mathbf{W}^{n+1})=\frac{1}{I_n I_{n+1}}\sum^{I_{n}}_{i=1} \sum^{I_{n+1}}_{j=1} W^{n+1}_{ij}$. 
The square of the last we denote as $\mu^2(\cdot)$. 
Vector $\mathbf{x}_{n}$ has some averaged expected value $\mu\big(E[\mathbf{x}_n]\big)$ and mean variance $\mu\big(\text{Var}[\mathbf{x}_n]\big)$. 
Again, both can be moved outside the sum, and therefore
\begin{equation}\label{Var_x_tilde3}
	\begin{array}{c}
		\text{Var}(\tilde{x}^t_{n+1,i}) \approx 2D^C_M I^2_n\mu^2(\mathbf{W}^{n+1})\mu^2(\text{E}[\mathbf{x}_n])  + \\
		2D^U_A I_n\eta(\mathbf{W}^{n+1}) + 2D^C_A I^2_n\mu^2(\mathbf{W}^{n+1}) + \\
		(1+2D^C_M)\mu^2(\text{E}[\mathbf{x}_n])\times \text{Var}\Big[ \sum\limits^{I_n}_{j=1} W^{n+1}_{ij} \times \\
		(1+\sqrt{2D^U_M}\xi^{U,M}_{nj}) \Big]  + \mu\big( \text{Var}[\mathbf{x}_n]\big)(1+2D^C_M)\times \\ 
		\Big( \big( \sum\limits^{I_n}_{j=1} W^{n+1}_{ij} \big)^2 + \sum\limits^{I_n}_{j=1}2D^U_M(W^{n+1}_{ij})^2  \Big) = \\
		I^2_n\mu^2(\mathbf{W}^{n+1})\cdot\Big( 2D^C_A+2D^C_M\big(\mu(\text{E}[\mathbf{x}_n])\big)^2 \Big) + \\
		I_n\eta(\mathbf{W}^{n+1})\cdot\Big( 2D^U_M(1+2D^C_M)\big(\mu(\text{E}[\mathbf{x}_n])\big)^2 +\\
		2D^U_A\Big) + I^2_n\mu^2(\mathbf{W}^{n+1})\Big(1+\frac{\eta(\mathbf{W}^{n+1})}{I_n\mu(\mathbf{W}^{n+1})}2D^U_M\Big)\times\\
		(1+2D^C_M)\mu(\text{Var}[\mathbf{x}_n])\big)  .
	\end{array}
\end{equation}

In the case of symmetric DNNs, $I^2_n\mu(\mathbf{W}^{n+1})=1$ and $I_n\eta(\mathbf{W}^{n+1})=1/I_n$.
When $I_n\gg 1$, all terms with $I_n\eta(\mathbf{W}^{n+1})$ can be neglected.
\begin{equation}\label{Var_x_tilde_hidden}
	\begin{array}{c}
		\text{Var}[\tilde{x}^t_{n+1,i}] \approx 2D^C_A + 2D^C_M\mu^2\big(E(\mathbf{x}^t_n)\big) +\\
		(1+2D^C_M)\mu\big( \text{Var}[\mathbf{x}^t_n] \big).
	\end{array}
\end{equation}
If $I_n=1$ , as in the first layer, then $I^2_n\mu(\mathbf{W}^{n+1})=I_n\eta(\mathbf{W}^{n+1})$ and the variance this layer becomes 
\begin{equation}\label{Var_x_tilde_first}
	\begin{array}{c}
		\text{Var}[\tilde{x}^t_{n+1,i}] = \sigma^2_+ + \sigma^2_\times\big(E[x^t_n]\big)^2 + \\
		(1+\sigma^2_\times)\mu\big( \text{Var}[{x}^t_n] \big).
	\end{array}
\end{equation}
Thus, the general equations of variance in symmetric DNNs are
\begin{equation}\label{Var_x_tilde_main}
	\begin{array}{c}
		\text{Var}[\tilde{x}^t_1] = \text{Var}[x^t_1] = 0 \\
		\text{Var}[\tilde{x}^t_{2,i}] = \sigma^2_+ + \sigma^2_\times\big(E[x^t_1]\big)^2.
	\end{array}
\end{equation}
The corresponding variance in hidden and output layers is described by Eq.~(\ref{Var_x_tilde_hidden}). The final output variance from any layer can be calculated using Eq.~(\ref{eq:EVar_one}).

~

\noindent \textbf{Higher order approximation of operator $\mathbf{\hat{F}}$.}

Operator $\mathbf{\hat{F}}$, introduced in Sect.~\ref{sec:sec2_high}, describes the variance of some random variable $Y$ which is equal to nonlinear function of another random variable: $Y=f(X)$. 
In general, variance is the difference between the expectation value of a squared variable and the square of expectation value. 
In order to calculate expectation values, nonlinear function $f(\cdot)$ has to be approximated, for which we use its Taylor series

\begin{equation}\label{eq:Taylor}
	\begin{array}{c}
		f(x)\approx\sum\limits^{M}_{m=0} \frac{1}{m!}f^{(m)}(\mu)(x-\mu)^m
		\approx \\
		\sum\limits^M_{m=0}T_m(x-\mu)^m 
		\approx f(\mu)+\sum\limits^M_{m=1}T_m(x-\mu)^m,
	\end{array}
\end{equation}
where $f^{(m)}$ denotes the $m$th derivative of function $f(\cdot)$, $T_m=\frac{1}{m!}f^{(m)}(\mu)$. If variable $X$ has the probability density distribution $p_X(x)$, then the expected value of $Y$ is
\begin{equation}\label{eq:E_Y}
	\begin{array}{c}
		\text{E}[Y]=\int\limits^{+\infty}_{-\infty} f(x)p_X(x)dx \approx
		\int\limits^{+\infty}_{-\infty} f(\mu)p_X(x)dx + \\ \int\limits^{+\infty}_{-\infty} \Big( \sum\limits^M_{m=1}T_m (x-\mu)^m \Big) p_X(x)dx = \\
		f(\mu) + \sum\limits^M_{m=1}T_m V_m,
	\end{array}
\end{equation}
where $V_m=\int\limits^{+\infty}_{-\infty} (x-\mu)^m p_X(x)dx$. The expected value of the square can be calculated in a similar way:
\begin{equation}\label{eq:E_Y2}
	\begin{array}{c}
		\text{E}[Y^2]\approx \int\limits^{+\infty}_{-\infty}\Big( f(\mu) +\sum\limits^M_{m=1}T_mV_m \Big)^2 p_X(x) dx = \\
		\big(f(\mu)\big)^2 + 2f(\mu)\sum\limits^M_{m=1}T_mV_m + \sum\limits^M_{m=1}T^2_mV_{2m} + \\
		2\sum\limits^M_{m=1}\sum\limits^{m-1}_{i=1}T_mT_iV_{m+i}.
	\end{array}
\end{equation}
Using the square of Eqs.~(\ref{eq:E_Y},\ref{eq:E_Y2}), the variance of the $M$th order is
\begin{equation}\label{eq:Var_Y_M}
	\begin{array}{c}
		\text{Var}[Y]\Big|_M = \text{E}[Y^2]-\big( \text{E}[Y] \big)^2 = \sum\limits^M_{m=1}T^2_m V_{2m} + \\
		2\sum\limits^M_{m=1}\sum\limits^{m-1}_{i=1} T_mT_iV_{m+i} - \big(  \sum\limits^M_{m=1} T_mV_m  \big)^2,
	\end{array}
\end{equation}
which can simply be rearranged into Eq.~(\ref{eq:F_M}). For some cases it would be more convenient to use the recurrent equation based on the approximation of the previous order:
\begin{equation}\label{eq:Var_Y_rec}
	\begin{array}{c}
		\text{Var}[Y]\Big|_M =2\sum\limits^{M-1}_{m=1}\sum\limits^{m-1}_{i=1} T_mT_iV_{m+i}+ T^2_MV_{2M} + \\
		\sum\limits^{M-1}_{m=1}T^2_m V_{2m} - \big( \sum\limits^{M-1}_{m=1} T_mV_m \big)^2 - \big(  \sum\limits^{M-1}_{m=1} T_mV_m  \big)^2 -\\
		T^2_mV^2_m = \text{Var}[Y]\Big|_{M-1} + T^2_M(V_{2M}-V^2_M) - \\
		2T_M\sum\limits^{M-1}_{i=1} T_i(V_{M+i}-V_MV_i),
	\end{array}
\end{equation}
which again can be rearranged into Eq.~(\ref{eq:F_M_rec}).

As an example, let us consider the cubic nonlinear function $f(x)=3x^2 - 2x^3$, which reassembles a sigmoid function in the range $(0;1)$. 
This nonlinearity has no parameters, and its second derivative is always larger one near $x=0.5$. 
The first order approximation based on Eq.~(\ref{eq:F_simple}) therefore leads to a growing error for deep networks, as was observed in Sect.~\ref{sec:SymDNN} for sigmoid function with $\alpha>2$. 

The new nonlinearity is cubic, so the maximal order of Taylor series approximation to precisely reproduce this function is $M=3$.
The Taylor series terms are $T_1=6(\mu-\mu^2)$, $T_2=3(1-2\mu)$, $T_3=-2$, and when $\mu=0.5$ they transform to $T_1=1.5$, $T_2=0$, $T_3=-2$.
Calculating the required $V$-values is less straight forward.
For any random $X$ it always holds that $V_1=0$ and $V_2=\text{Var}[X]$. 
The following terms strongly depend on the probability distribution of $X$, and can be calculated only if $p_X(x)$ is known or can be well approximated.
For example, if $X$ has a normal distribution, then $V_3=V_5=0$, $V_4=3\big(\text{Var}[X]\big)^2$, $V_6=15\big(\text{Var}[X]\big)^3$.

\subsection{Approximation of state probability $p^*(\cdot)$}\label{sec:Methods_pdf}

In order to approximate the impact of noise in trained DNNs we need to take the influence of bias and connection matrices into account, which were introduced in Eq.~(\ref{eq:mu_eta_integrals}). 
The probability density distribution $p^*_n(z)$ of $\tilde{\mathbf{x}}_n$ takes part in all terms of Eq.~(\ref{eq:mu_eta_integrals}).

\begin{figure}[h]
	\center{\includegraphics[width=1\linewidth]{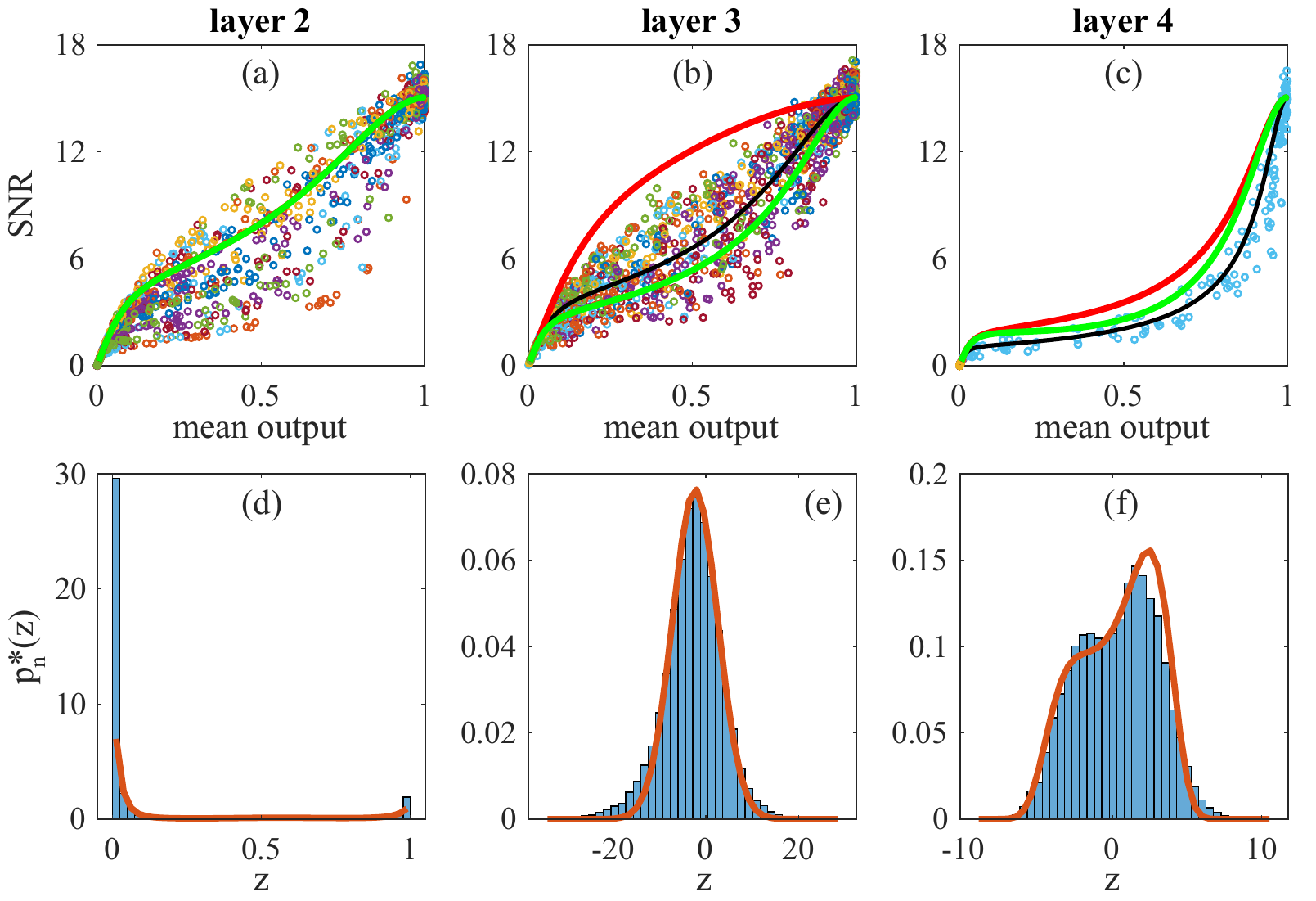}}
	\caption[]{SNR in hidden and output layers of Network A, used for digits recognition are given in panels (a,b,c). Red, green and black lines correspond to analytical prediction based on assumption of uniform, normal and using $g(\tilde{x})$ for approximating
		$p^*_n(z)$, respectively. The probability distribution of $\tilde{\mathbf{x}}_n$ in layers 2,3,4 and corresponding approximating curves are shown in panels (d--f).}
	\label{fig:approximation_p}
\end{figure}

One can make the weak assumption that, for DNN's with many neurons in its hidden layers, $p^*_n(z)$ to be some well behaved function, such as uniform or normal distributions. 
Figure~\ref{fig:approximation_p}(a--c) shows the SNR for DNN A, based on weight statistics using uniform (red line) and normal (green line) distributions.

Reaching a quantitative agreement based on the statistics of weights requires approximating function $p^*_n(z)$.
Here, we use 
\begin{equation}\label{eq:StateApprox}
	g(x)=\exp(c_4 x^4+c_3 x^3+c_2 x^2+c_1 x+c_0)
\end{equation}
\noindent for our approximation.
It allows approximating bimodal, normal and uniform distributions and hence captures a wide range of cases which excellently approximate all conditions we observed for DNNs trained for classifying as well as for continuous function approximation. 
The analytical prediction of SNR curve based on this approximation is given in Fig.~\ref{fig:approximation_p}(a--c) by black line. 
As can be seen from Fig.~\ref{fig:approximation_p}(d--f) the obtained analytical prediction is in a good agreement with the statistics of $\tilde{\mathbf{x}}_n$ in each layer.

Analytics based on the approximated $p^*_n(z)$ via $g(\tilde{x})$ demonstrated the best agreement with numerical simulation. 
At the same time, uniform and normal distributions do to some degree produce similar behaviour. 
They do not demonstrate such good quantitative agreement.


\bibliography{references}

\begin{thebibliography}{43}%
\makeatletter
\providecommand \@ifxundefined [1]{%
 \@ifx{#1\undefined}
}%
\providecommand \@ifnum [1]{%
 \ifnum #1\expandafter \@firstoftwo
 \else \expandafter \@secondoftwo
 \fi
}%
\providecommand \@ifx [1]{%
 \ifx #1\expandafter \@firstoftwo
 \else \expandafter \@secondoftwo
 \fi
}%
\providecommand \natexlab [1]{#1}%
\providecommand \enquote  [1]{``#1''}%
\providecommand \bibnamefont  [1]{#1}%
\providecommand \bibfnamefont [1]{#1}%
\providecommand \citenamefont [1]{#1}%
\providecommand \href@noop [0]{\@secondoftwo}%
\providecommand \href [0]{\begingroup \@sanitize@url \@href}%
\providecommand \@href[1]{\@@startlink{#1}\@@href}%
\providecommand \@@href[1]{\endgroup#1\@@endlink}%
\providecommand \@sanitize@url [0]{\catcode `\\12\catcode `\$12\catcode
  `\&12\catcode `\#12\catcode `\^12\catcode `\_12\catcode `\%12\relax}%
\providecommand \@@startlink[1]{}%
\providecommand \@@endlink[0]{}%
\providecommand \url  [0]{\begingroup\@sanitize@url \@url }%
\providecommand \@url [1]{\endgroup\@href {#1}{\urlprefix }}%
\providecommand \urlprefix  [0]{URL }%
\providecommand \Eprint [0]{\href }%
\providecommand \doibase [0]{http://dx.doi.org/}%
\providecommand \selectlanguage [0]{\@gobble}%
\providecommand \bibinfo  [0]{\@secondoftwo}%
\providecommand \bibfield  [0]{\@secondoftwo}%
\providecommand \translation [1]{[#1]}%
\providecommand \BibitemOpen [0]{}%
\providecommand \bibitemStop [0]{}%
\providecommand \bibitemNoStop [0]{.\EOS\space}%
\providecommand \EOS [0]{\spacefactor3000\relax}%
\providecommand \BibitemShut  [1]{\csname bibitem#1\endcsname}%
\let\auto@bib@innerbib\@empty
\bibitem [{\citenamefont {Boahen}(2017)}]{Boahen2017}%
  \BibitemOpen
  \bibfield  {author} {\bibinfo {author} {\bibfnamefont {K.}~\bibnamefont
  {Boahen}},\ }\href {\doibase 10.1109/MCSE.2017.33} {\bibfield  {journal}
  {\bibinfo  {journal} {Computing in Science \& Engineering}\ }\textbf
  {\bibinfo {volume} {19}},\ \bibinfo {pages} {14} (\bibinfo {year}
  {2017})}\BibitemShut {NoStop}%
\bibitem [{\citenamefont {Gupta}\ \emph {et~al.}(2015)\citenamefont {Gupta},
  \citenamefont {Agrawal}, \citenamefont {Gopalakrishnan},\ and\ \citenamefont
  {Narayanan}}]{Gupta2015}%
  \BibitemOpen
  \bibfield  {author} {\bibinfo {author} {\bibfnamefont {S.}~\bibnamefont
  {Gupta}}, \bibinfo {author} {\bibfnamefont {A.}~\bibnamefont {Agrawal}},
  \bibinfo {author} {\bibfnamefont {K.}~\bibnamefont {Gopalakrishnan}}, \ and\
  \bibinfo {author} {\bibfnamefont {P.}~\bibnamefont {Narayanan}},\ }\href
  {\doibase 10.1109/72.80206} {\bibfield  {journal} {\bibinfo  {journal}
  {Proceedings of the 32nd International Conference on International Conference
  on Machine Learning}\ }\textbf {\bibinfo {volume} {37}},\ \bibinfo {pages}
  {1737} (\bibinfo {year} {2015})}\BibitemShut {NoStop}%
\bibitem [{\citenamefont {Hasler}\ and\ \citenamefont
  {Marr}(2013)}]{Hasler2013}%
  \BibitemOpen
  \bibfield  {author} {\bibinfo {author} {\bibfnamefont {J.}~\bibnamefont
  {Hasler}}\ and\ \bibinfo {author} {\bibfnamefont {B.}~\bibnamefont {Marr}},\
  }\href {\doibase 10.3389/fnins.2013.00118} {\bibfield  {journal} {\bibinfo
  {journal} {Frontiers in neuroscience}\ }\textbf {\bibinfo {volume} {7}},\
  \bibinfo {pages} {118} (\bibinfo {year} {2013})}\BibitemShut {NoStop}%
\bibitem [{\citenamefont {Brunner}\ \emph {et~al.}(2013)\citenamefont
  {Brunner}, \citenamefont {Soriano}, \citenamefont {Mirasso},\ and\
  \citenamefont {Fischer}}]{Brunner2013a}%
  \BibitemOpen
  \bibfield  {author} {\bibinfo {author} {\bibfnamefont {D.}~\bibnamefont
  {Brunner}}, \bibinfo {author} {\bibfnamefont {M.~C.}\ \bibnamefont
  {Soriano}}, \bibinfo {author} {\bibfnamefont {C.~R.}\ \bibnamefont
  {Mirasso}}, \ and\ \bibinfo {author} {\bibfnamefont {I.}~\bibnamefont
  {Fischer}},\ }\href@noop {} {\bibfield  {journal} {\bibinfo  {journal}
  {Nature communications}\ }\textbf {\bibinfo {volume} {4}},\ \bibinfo {pages}
  {1364} (\bibinfo {year} {2013})}\BibitemShut {NoStop}%
\bibitem [{\citenamefont {Tuma}\ \emph {et~al.}(2016)\citenamefont {Tuma},
  \citenamefont {Pantazi}, \citenamefont {{Le Gallo}}, \citenamefont
  {Sebastian},\ and\ \citenamefont {Eleftheriou}}]{Tuma2016}%
  \BibitemOpen
  \bibfield  {author} {\bibinfo {author} {\bibfnamefont {T.}~\bibnamefont
  {Tuma}}, \bibinfo {author} {\bibfnamefont {A.}~\bibnamefont {Pantazi}},
  \bibinfo {author} {\bibfnamefont {M.}~\bibnamefont {{Le Gallo}}}, \bibinfo
  {author} {\bibfnamefont {A.}~\bibnamefont {Sebastian}}, \ and\ \bibinfo
  {author} {\bibfnamefont {E.}~\bibnamefont {Eleftheriou}},\ }\href@noop {}
  {\bibfield  {journal} {\bibinfo  {journal} {Nature Nanotechnology}\ }\textbf
  {\bibinfo {volume} {11}},\ \bibinfo {pages} {693} (\bibinfo {year}
  {2016})}\BibitemShut {NoStop}%
\bibitem [{Tor(2017)}]{Torrejon2017}%
  \BibitemOpen
  \href@noop {} {\bibfield  {journal} {\bibinfo  {journal} {Nature}\ }\textbf
  {\bibinfo {volume} {547}},\ \bibinfo {pages} {428} (\bibinfo {year}
  {2017})}\BibitemShut {NoStop}%
\bibitem [{\citenamefont {Psaltis}\ \emph {et~al.}(1990)\citenamefont
  {Psaltis}, \citenamefont {Brady}, \citenamefont {Gu},\ and\ \citenamefont
  {Lin}}]{Psaltis1990}%
  \BibitemOpen
  \bibfield  {author} {\bibinfo {author} {\bibfnamefont {D.}~\bibnamefont
  {Psaltis}}, \bibinfo {author} {\bibfnamefont {D.}~\bibnamefont {Brady}},
  \bibinfo {author} {\bibfnamefont {X.-G.}\ \bibnamefont {Gu}}, \ and\ \bibinfo
  {author} {\bibfnamefont {S.}~\bibnamefont {Lin}},\ }\href {\doibase
  10.1038/343325a0} {\bibfield  {journal} {\bibinfo  {journal} {Nature}\
  }\textbf {\bibinfo {volume} {343}},\ \bibinfo {pages} {325} (\bibinfo {year}
  {1990})}\BibitemShut {NoStop}%
\bibitem [{\citenamefont {Bueno}\ \emph {et~al.}(2018)\citenamefont {Bueno},
  \citenamefont {Maktoobi}, \citenamefont {Froehly}, \citenamefont {Fischer},
  \citenamefont {Jacquot}, \citenamefont {Larger},\ and\ \citenamefont
  {Brunner}}]{Bueno2018}%
  \BibitemOpen
  \bibfield  {author} {\bibinfo {author} {\bibfnamefont {J.}~\bibnamefont
  {Bueno}}, \bibinfo {author} {\bibfnamefont {S.}~\bibnamefont {Maktoobi}},
  \bibinfo {author} {\bibfnamefont {L.}~\bibnamefont {Froehly}}, \bibinfo
  {author} {\bibfnamefont {I.}~\bibnamefont {Fischer}}, \bibinfo {author}
  {\bibfnamefont {M.}~\bibnamefont {Jacquot}}, \bibinfo {author} {\bibfnamefont
  {L.}~\bibnamefont {Larger}}, \ and\ \bibinfo {author} {\bibfnamefont
  {D.}~\bibnamefont {Brunner}},\ }\href@noop {} {\bibfield  {journal} {\bibinfo
   {journal} {Optica}\ }\textbf {\bibinfo {volume} {5}},\ \bibinfo {pages} {756
  } (\bibinfo {year} {2018})}\BibitemShut {NoStop}%
\bibitem [{\citenamefont {Lin}\ \emph {et~al.}(2018)\citenamefont {Lin},
  \citenamefont {Rivenson}, \citenamefont {Yardimci}, \citenamefont {Veli},
  \citenamefont {Jarrahi},\ and\ \citenamefont {Ozcan}}]{Lin2018}%
  \BibitemOpen
  \bibfield  {author} {\bibinfo {author} {\bibfnamefont {X.}~\bibnamefont
  {Lin}}, \bibinfo {author} {\bibfnamefont {Y.}~\bibnamefont {Rivenson}},
  \bibinfo {author} {\bibfnamefont {N.~T.}\ \bibnamefont {Yardimci}}, \bibinfo
  {author} {\bibfnamefont {M.}~\bibnamefont {Veli}}, \bibinfo {author}
  {\bibfnamefont {M.}~\bibnamefont {Jarrahi}}, \ and\ \bibinfo {author}
  {\bibfnamefont {A.}~\bibnamefont {Ozcan}},\ }\href {\doibase
  10.1126/science.aat8084} {\bibfield  {journal} {\bibinfo  {journal}
  {Science}\ }\textbf {\bibinfo {volume} {26}},\ \bibinfo {pages} {1} (\bibinfo
  {year} {2018})}\BibitemShut {NoStop}%
\bibitem [{\citenamefont {Shen}\ \emph {et~al.}(2017)\citenamefont {Shen},
  \citenamefont {Harris}, \citenamefont {Skirlo}, \citenamefont {Prabhu},
  \citenamefont {Baehr-Jones}, \citenamefont {Hochberg}, \citenamefont {Sun},
  \citenamefont {Zhao}, \citenamefont {Larochelle}, \citenamefont {Englund},\
  and\ \citenamefont {Soljacic}}]{Shen2016}%
  \BibitemOpen
  \bibfield  {author} {\bibinfo {author} {\bibfnamefont {Y.}~\bibnamefont
  {Shen}}, \bibinfo {author} {\bibfnamefont {N.~C.}\ \bibnamefont {Harris}},
  \bibinfo {author} {\bibfnamefont {S.}~\bibnamefont {Skirlo}}, \bibinfo
  {author} {\bibfnamefont {M.}~\bibnamefont {Prabhu}}, \bibinfo {author}
  {\bibfnamefont {T.}~\bibnamefont {Baehr-Jones}}, \bibinfo {author}
  {\bibfnamefont {M.}~\bibnamefont {Hochberg}}, \bibinfo {author}
  {\bibfnamefont {X.}~\bibnamefont {Sun}}, \bibinfo {author} {\bibfnamefont
  {S.}~\bibnamefont {Zhao}}, \bibinfo {author} {\bibfnamefont {H.}~\bibnamefont
  {Larochelle}}, \bibinfo {author} {\bibfnamefont {D.}~\bibnamefont {Englund}},
  \ and\ \bibinfo {author} {\bibfnamefont {M.}~\bibnamefont {Soljacic}},\
  }\href@noop {} {\bibfield  {journal} {\bibinfo  {journal} {Nature Photonics}\
  }\textbf {\bibinfo {volume} {11}},\ \bibinfo {pages} {441} (\bibinfo {year}
  {2017})}\BibitemShut {NoStop}%
\bibitem [{\citenamefont {Tait}\ \emph {et~al.}(2017)\citenamefont {Tait},
  \citenamefont {{De Lima}}, \citenamefont {Zhou}, \citenamefont {Wu},
  \citenamefont {Nahmias}, \citenamefont {Shastri},\ and\ \citenamefont
  {Prucnal}}]{Tait2017}%
  \BibitemOpen
  \bibfield  {author} {\bibinfo {author} {\bibfnamefont {A.~N.}\ \bibnamefont
  {Tait}}, \bibinfo {author} {\bibfnamefont {T.~F.}\ \bibnamefont {{De Lima}}},
  \bibinfo {author} {\bibfnamefont {E.}~\bibnamefont {Zhou}}, \bibinfo {author}
  {\bibfnamefont {A.~X.}\ \bibnamefont {Wu}}, \bibinfo {author} {\bibfnamefont
  {M.~A.}\ \bibnamefont {Nahmias}}, \bibinfo {author} {\bibfnamefont {B.~J.}\
  \bibnamefont {Shastri}}, \ and\ \bibinfo {author} {\bibfnamefont {P.~R.}\
  \bibnamefont {Prucnal}},\ }\href@noop {} {\bibfield  {journal} {\bibinfo
  {journal} {Scientific Reports}\ }\textbf {\bibinfo {volume} {7}},\ \bibinfo
  {pages} {1} (\bibinfo {year} {2017})}\BibitemShut {NoStop}%
\bibitem [{\citenamefont {Moughames}\ \emph
  {et~al.}(2020{\natexlab{a}})\citenamefont {Moughames}, \citenamefont {Porte},
  \citenamefont {Thiel}, \citenamefont {Ulliac}, \citenamefont {Larger},
  \citenamefont {Jacquot}, \citenamefont {Kadic},\ and\ \citenamefont
  {Brunner}}]{Moughames2020}%
  \BibitemOpen
  \bibfield  {author} {\bibinfo {author} {\bibfnamefont {J.}~\bibnamefont
  {Moughames}}, \bibinfo {author} {\bibfnamefont {X.}~\bibnamefont {Porte}},
  \bibinfo {author} {\bibfnamefont {M.}~\bibnamefont {Thiel}}, \bibinfo
  {author} {\bibfnamefont {G.}~\bibnamefont {Ulliac}}, \bibinfo {author}
  {\bibfnamefont {L.}~\bibnamefont {Larger}}, \bibinfo {author} {\bibfnamefont
  {M.}~\bibnamefont {Jacquot}}, \bibinfo {author} {\bibfnamefont
  {M.}~\bibnamefont {Kadic}}, \ and\ \bibinfo {author} {\bibfnamefont
  {D.}~\bibnamefont {Brunner}},\ }\href {\doibase 10.1364/OPTICA.388205}
  {\bibfield  {journal} {\bibinfo  {journal} {Optica}\ }\textbf {\bibinfo
  {volume} {7}},\ \bibinfo {pages} {640} (\bibinfo {year}
  {2020}{\natexlab{a}})}\BibitemShut {NoStop}%
\bibitem [{\citenamefont {{Dinc, Niyazi Ulas}}, \citenamefont {{Psaltis,
  Demetri}},\ and\ \citenamefont {{Brunner, Daniel}}(2020)}]{Dinc2020}%
  \BibitemOpen
  \bibfield  {author} {\bibinfo {author} {\bibnamefont {{Dinc, Niyazi Ulas}}},
  \bibinfo {author} {\bibnamefont {{Psaltis, Demetri}}}, \ and\ \bibinfo
  {author} {\bibnamefont {{Brunner, Daniel}}},\ }\href {\doibase
  10.1051/photon/202010434} {\bibfield  {journal} {\bibinfo  {journal}
  {Photoniques}\ ,\ \bibinfo {pages} {34}} (\bibinfo {year}
  {2020})}\BibitemShut {NoStop}%
\bibitem [{\citenamefont {Moughames}\ \emph
  {et~al.}(2020{\natexlab{b}})\citenamefont {Moughames}, \citenamefont {Porte},
  \citenamefont {Larger}, \citenamefont {Jacquot}, \citenamefont {Kadic},\ and\
  \citenamefont {Brunner}}]{Moughames-2-2020}%
  \BibitemOpen
  \bibfield  {author} {\bibinfo {author} {\bibfnamefont {J.}~\bibnamefont
  {Moughames}}, \bibinfo {author} {\bibfnamefont {X.}~\bibnamefont {Porte}},
  \bibinfo {author} {\bibfnamefont {L.}~\bibnamefont {Larger}}, \bibinfo
  {author} {\bibfnamefont {M.}~\bibnamefont {Jacquot}}, \bibinfo {author}
  {\bibfnamefont {M.}~\bibnamefont {Kadic}}, \ and\ \bibinfo {author}
  {\bibfnamefont {D.}~\bibnamefont {Brunner}},\ }\href {\doibase
  10.1364/OME.402974} {\bibfield  {journal} {\bibinfo  {journal} {Opt. Mater.
  Express}\ }\textbf {\bibinfo {volume} {10}},\ \bibinfo {pages} {2952}
  (\bibinfo {year} {2020}{\natexlab{b}})}\BibitemShut {NoStop}%
\bibitem [{\citenamefont {Wang}\ \emph {et~al.}(2018)\citenamefont {Wang},
  \citenamefont {Joshi}, \citenamefont {Savel'Ev}, \citenamefont {Song},
  \citenamefont {Midya}, \citenamefont {Li}, \citenamefont {Rao}, \citenamefont
  {Yan}, \citenamefont {Asapu}, \citenamefont {Zhuo}, \citenamefont {Jiang},
  \citenamefont {Lin}, \citenamefont {Li}, \citenamefont {Yoon}, \citenamefont
  {Upadhyay}, \citenamefont {Zhang}, \citenamefont {Hu}, \citenamefont
  {Strachan}, \citenamefont {Barnell}, \citenamefont {Wu}, \citenamefont {Wu},
  \citenamefont {Williams}, \citenamefont {Xia},\ and\ \citenamefont
  {Yang}}]{Wang2018}%
  \BibitemOpen
  \bibfield  {author} {\bibinfo {author} {\bibfnamefont {Z.}~\bibnamefont
  {Wang}}, \bibinfo {author} {\bibfnamefont {S.}~\bibnamefont {Joshi}},
  \bibinfo {author} {\bibfnamefont {S.}~\bibnamefont {Savel'Ev}}, \bibinfo
  {author} {\bibfnamefont {W.}~\bibnamefont {Song}}, \bibinfo {author}
  {\bibfnamefont {R.}~\bibnamefont {Midya}}, \bibinfo {author} {\bibfnamefont
  {Y.}~\bibnamefont {Li}}, \bibinfo {author} {\bibfnamefont {M.}~\bibnamefont
  {Rao}}, \bibinfo {author} {\bibfnamefont {P.}~\bibnamefont {Yan}}, \bibinfo
  {author} {\bibfnamefont {S.}~\bibnamefont {Asapu}}, \bibinfo {author}
  {\bibfnamefont {Y.}~\bibnamefont {Zhuo}}, \bibinfo {author} {\bibfnamefont
  {H.}~\bibnamefont {Jiang}}, \bibinfo {author} {\bibfnamefont
  {P.}~\bibnamefont {Lin}}, \bibinfo {author} {\bibfnamefont {C.}~\bibnamefont
  {Li}}, \bibinfo {author} {\bibfnamefont {J.~H.}\ \bibnamefont {Yoon}},
  \bibinfo {author} {\bibfnamefont {N.~K.}\ \bibnamefont {Upadhyay}}, \bibinfo
  {author} {\bibfnamefont {J.}~\bibnamefont {Zhang}}, \bibinfo {author}
  {\bibfnamefont {M.}~\bibnamefont {Hu}}, \bibinfo {author} {\bibfnamefont
  {J.~P.}\ \bibnamefont {Strachan}}, \bibinfo {author} {\bibfnamefont
  {M.}~\bibnamefont {Barnell}}, \bibinfo {author} {\bibfnamefont
  {Q.}~\bibnamefont {Wu}}, \bibinfo {author} {\bibfnamefont {H.}~\bibnamefont
  {Wu}}, \bibinfo {author} {\bibfnamefont {R.~S.}\ \bibnamefont {Williams}},
  \bibinfo {author} {\bibfnamefont {Q.}~\bibnamefont {Xia}}, \ and\ \bibinfo
  {author} {\bibfnamefont {J.~J.}\ \bibnamefont {Yang}},\ }\href {\doibase
  10.1038/s41928-018-0023-2} {\bibfield  {journal} {\bibinfo  {journal} {Nature
  Electronics}\ }\textbf {\bibinfo {volume} {1}},\ \bibinfo {pages} {137}
  (\bibinfo {year} {2018})}\BibitemShut {NoStop}%
\bibitem [{\citenamefont {Lin}\ \emph {et~al.}(2020)\citenamefont {Lin},
  \citenamefont {Li}, \citenamefont {Wang}, \citenamefont {Li}, \citenamefont
  {Jiang}, \citenamefont {Song}, \citenamefont {Rao}, \citenamefont {Zhuo},
  \citenamefont {Upadhyay}, \citenamefont {Barnell}, \citenamefont {Wu},
  \citenamefont {Yang},\ and\ \citenamefont {Xia}}]{Lin2020}%
  \BibitemOpen
  \bibfield  {author} {\bibinfo {author} {\bibfnamefont {P.}~\bibnamefont
  {Lin}}, \bibinfo {author} {\bibfnamefont {C.}~\bibnamefont {Li}}, \bibinfo
  {author} {\bibfnamefont {Z.}~\bibnamefont {Wang}}, \bibinfo {author}
  {\bibfnamefont {Y.}~\bibnamefont {Li}}, \bibinfo {author} {\bibfnamefont
  {H.}~\bibnamefont {Jiang}}, \bibinfo {author} {\bibfnamefont
  {W.}~\bibnamefont {Song}}, \bibinfo {author} {\bibfnamefont {M.}~\bibnamefont
  {Rao}}, \bibinfo {author} {\bibfnamefont {Y.}~\bibnamefont {Zhuo}}, \bibinfo
  {author} {\bibfnamefont {N.~K.}\ \bibnamefont {Upadhyay}}, \bibinfo {author}
  {\bibfnamefont {M.}~\bibnamefont {Barnell}}, \bibinfo {author} {\bibfnamefont
  {Q.}~\bibnamefont {Wu}}, \bibinfo {author} {\bibfnamefont {J.~J.}\
  \bibnamefont {Yang}}, \ and\ \bibinfo {author} {\bibfnamefont
  {Q.}~\bibnamefont {Xia}},\ }\href {\doibase 10.1038/s41928-020-0397-9}
  {\bibfield  {journal} {\bibinfo  {journal} {Nature Electronics}\ }\textbf
  {\bibinfo {volume} {3}},\ \bibinfo {pages} {225} (\bibinfo {year}
  {2020})}\BibitemShut {NoStop}%
\bibitem [{\citenamefont {Xia}\ and\ \citenamefont {Yang}(2019)}]{Xia2019}%
  \BibitemOpen
  \bibfield  {author} {\bibinfo {author} {\bibfnamefont {Q.}~\bibnamefont
  {Xia}}\ and\ \bibinfo {author} {\bibfnamefont {J.~J.}\ \bibnamefont {Yang}},\
  }\href {\doibase 10.1038/s41563-019-0291-x} {\bibfield  {journal} {\bibinfo
  {journal} {Nature Materials}\ }\textbf {\bibinfo {volume} {18}},\ \bibinfo
  {pages} {309} (\bibinfo {year} {2019})}\BibitemShut {NoStop}%
\bibitem [{\citenamefont {Feldmann}\ \emph {et~al.}(2021)\citenamefont
  {Feldmann}, \citenamefont {Youngblood}, \citenamefont {Karpov}, \citenamefont
  {Gehring}, \citenamefont {Li}, \citenamefont {Stappers}, \citenamefont
  {Le~Gallo}, \citenamefont {Fu}, \citenamefont {Lukashchuk}, \citenamefont
  {Raja}, \citenamefont {Liu}, \citenamefont {Wright}, \citenamefont
  {Sebastian}, \citenamefont {Kippenberg}, \citenamefont {Pernice},\ and\
  \citenamefont {Bhaskaran}}]{Feldmann2021}%
  \BibitemOpen
  \bibfield  {author} {\bibinfo {author} {\bibfnamefont {J.}~\bibnamefont
  {Feldmann}}, \bibinfo {author} {\bibfnamefont {N.}~\bibnamefont
  {Youngblood}}, \bibinfo {author} {\bibfnamefont {M.}~\bibnamefont {Karpov}},
  \bibinfo {author} {\bibfnamefont {H.}~\bibnamefont {Gehring}}, \bibinfo
  {author} {\bibfnamefont {X.}~\bibnamefont {Li}}, \bibinfo {author}
  {\bibfnamefont {M.}~\bibnamefont {Stappers}}, \bibinfo {author}
  {\bibfnamefont {M.}~\bibnamefont {Le~Gallo}}, \bibinfo {author}
  {\bibfnamefont {X.}~\bibnamefont {Fu}}, \bibinfo {author} {\bibfnamefont
  {A.}~\bibnamefont {Lukashchuk}}, \bibinfo {author} {\bibfnamefont {A.~S.}\
  \bibnamefont {Raja}}, \bibinfo {author} {\bibfnamefont {J.}~\bibnamefont
  {Liu}}, \bibinfo {author} {\bibfnamefont {C.~D.}\ \bibnamefont {Wright}},
  \bibinfo {author} {\bibfnamefont {A.}~\bibnamefont {Sebastian}}, \bibinfo
  {author} {\bibfnamefont {T.~J.}\ \bibnamefont {Kippenberg}}, \bibinfo
  {author} {\bibfnamefont {W.~H.~P.}\ \bibnamefont {Pernice}}, \ and\ \bibinfo
  {author} {\bibfnamefont {H.}~\bibnamefont {Bhaskaran}},\ }\href {\doibase
  10.1038/s41586-020-03070-1} {\bibfield  {journal} {\bibinfo  {journal}
  {Nature}\ }\textbf {\bibinfo {volume} {589}},\ \bibinfo {pages} {52}
  (\bibinfo {year} {2021})}\BibitemShut {NoStop}%
\bibitem [{\citenamefont {Moon}, \citenamefont {Shin},\ and\ \citenamefont
  {Jeon}(2019)}]{Moon2019}%
  \BibitemOpen
  \bibfield  {author} {\bibinfo {author} {\bibfnamefont {S.}~\bibnamefont
  {Moon}}, \bibinfo {author} {\bibfnamefont {K.}~\bibnamefont {Shin}}, \ and\
  \bibinfo {author} {\bibfnamefont {D.}~\bibnamefont {Jeon}},\ }\href {\doibase
  10.1109/TVLSI.2019.2893256} {\bibfield  {journal} {\bibinfo  {journal} {IEEE
  Transactions on Very Large Scale Integration (VLSI) Systems}\ }\textbf
  {\bibinfo {volume} {27}},\ \bibinfo {pages} {1455–1459} (\bibinfo {year}
  {2019})}\BibitemShut {NoStop}%
\bibitem [{\citenamefont {{Janke}}\ and\ \citenamefont
  {{Anderson}}(2020)}]{Janke2020}%
  \BibitemOpen
  \bibfield  {author} {\bibinfo {author} {\bibfnamefont {D.}~\bibnamefont
  {{Janke}}}\ and\ \bibinfo {author} {\bibfnamefont {D.~V.}\ \bibnamefont
  {{Anderson}}},\ }in\ \href {\doibase 10.1109/MWSCAS48704.2020.9184644} {\emph
  {\bibinfo {booktitle} {2020 IEEE 63rd International Midwest Symposium on
  Circuits and Systems (MWSCAS)}}}\ (\bibinfo {year} {2020})\ pp.\ \bibinfo
  {pages} {150--153}\BibitemShut {NoStop}%
\bibitem [{\citenamefont {Dolenko}\ and\ \citenamefont
  {Card}(1993)}]{Dolenko1993}%
  \BibitemOpen
  \bibfield  {author} {\bibinfo {author} {\bibfnamefont {B.}~\bibnamefont
  {Dolenko}}\ and\ \bibinfo {author} {\bibfnamefont {H.}~\bibnamefont {Card}},\
  }\href@noop {} {\bibfield  {journal} {\bibinfo  {journal} {Electronics
  letters}\ }\textbf {\bibinfo {volume} {29}},\ \bibinfo {pages} {693}
  (\bibinfo {year} {1993})}\BibitemShut {NoStop}%
\bibitem [{\citenamefont {Misra}\ and\ \citenamefont {Saha}(2010)}]{Misra2010}%
  \BibitemOpen
  \bibfield  {author} {\bibinfo {author} {\bibfnamefont {J.}~\bibnamefont
  {Misra}}\ and\ \bibinfo {author} {\bibfnamefont {I.}~\bibnamefont {Saha}},\
  }\href {\doibase https://doi.org/10.1016/j.neucom.2010.03.021} {\bibfield
  {journal} {\bibinfo  {journal} {Neurocomputing}\ }\textbf {\bibinfo {volume}
  {74}},\ \bibinfo {pages} {239} (\bibinfo {year} {2010})},\ \bibinfo {note}
  {artificial Brains}\BibitemShut {NoStop}%
\bibitem [{\citenamefont {{Dibazar}}\ \emph {et~al.}(2006)\citenamefont
  {{Dibazar}}, \citenamefont {{Bangalore}}, \citenamefont {{Hyungook Park}},
  \citenamefont {{George}}, \citenamefont {{Yamada}},\ and\ \citenamefont
  {{Berger}}}]{Dibazar2006}%
  \BibitemOpen
  \bibfield  {author} {\bibinfo {author} {\bibfnamefont {A.~A.}\ \bibnamefont
  {{Dibazar}}}, \bibinfo {author} {\bibfnamefont {A.}~\bibnamefont
  {{Bangalore}}}, \bibinfo {author} {\bibnamefont {{Hyungook Park}}}, \bibinfo
  {author} {\bibfnamefont {S.}~\bibnamefont {{George}}}, \bibinfo {author}
  {\bibfnamefont {W.}~\bibnamefont {{Yamada}}}, \ and\ \bibinfo {author}
  {\bibfnamefont {T.~W.}\ \bibnamefont {{Berger}}},\ }in\ \href {\doibase
  10.1109/IJCNN.2006.246949} {\emph {\bibinfo {booktitle} {The 2006 IEEE
  International Joint Conference on Neural Network Proceedings}}}\ (\bibinfo
  {year} {2006})\ pp.\ \bibinfo {pages} {2015--2022}\BibitemShut {NoStop}%
\bibitem [{\citenamefont {Soriano}\ \emph {et~al.}(2015)\citenamefont
  {Soriano}, \citenamefont {Ortín}, \citenamefont {Keuninckx}, \citenamefont
  {Appeltant}, \citenamefont {Danckaert}, \citenamefont {Pesquera},\ and\
  \citenamefont {van~der Sande}}]{Soriano2015}%
  \BibitemOpen
  \bibfield  {author} {\bibinfo {author} {\bibfnamefont {M.~C.}\ \bibnamefont
  {Soriano}}, \bibinfo {author} {\bibfnamefont {S.}~\bibnamefont {Ortín}},
  \bibinfo {author} {\bibfnamefont {L.}~\bibnamefont {Keuninckx}}, \bibinfo
  {author} {\bibfnamefont {L.}~\bibnamefont {Appeltant}}, \bibinfo {author}
  {\bibfnamefont {J.}~\bibnamefont {Danckaert}}, \bibinfo {author}
  {\bibfnamefont {L.}~\bibnamefont {Pesquera}}, \ and\ \bibinfo {author}
  {\bibfnamefont {G.}~\bibnamefont {van~der Sande}},\ }\href {\doibase
  https://doi.org/10.1109/TNNLS.2014.2311855} {\bibfield  {journal} {\bibinfo
  {journal} {IEEE transactions on neural networks and learning systems}\
  }\textbf {\bibinfo {volume} {26}},\ \bibinfo {pages} {388–393} (\bibinfo
  {year} {2015})}\BibitemShut {NoStop}%
\bibitem [{\citenamefont {Frye}, \citenamefont {Rietman},\ and\ \citenamefont
  {Wong}(1991)}]{Frye1991}%
  \BibitemOpen
  \bibfield  {author} {\bibinfo {author} {\bibfnamefont {R.}~\bibnamefont
  {Frye}}, \bibinfo {author} {\bibfnamefont {E.}~\bibnamefont {Rietman}}, \
  and\ \bibinfo {author} {\bibfnamefont {C.}~\bibnamefont {Wong}},\ }\href
  {\doibase 10.1109/72.80296} {\bibfield  {journal} {\bibinfo  {journal} {IEEE
  Transactions on Neural Networks}\ }\textbf {\bibinfo {volume} {2}},\ \bibinfo
  {pages} {110} (\bibinfo {year} {1991})}\BibitemShut {NoStop}%
\bibitem [{\citenamefont {Gailey}\ \emph {et~al.}(1997)\citenamefont {Gailey},
  \citenamefont {Neiman}, \citenamefont {Collins},\ and\ \citenamefont
  {Moss}}]{Gailey1997}%
  \BibitemOpen
  \bibfield  {author} {\bibinfo {author} {\bibfnamefont {P.~C.}\ \bibnamefont
  {Gailey}}, \bibinfo {author} {\bibfnamefont {A.}~\bibnamefont {Neiman}},
  \bibinfo {author} {\bibfnamefont {J.~J.}\ \bibnamefont {Collins}}, \ and\
  \bibinfo {author} {\bibfnamefont {F.}~\bibnamefont {Moss}},\ }\href {\doibase
  10.1103/PhysRevLett.79.4701} {\bibfield  {journal} {\bibinfo  {journal}
  {Phys. Rev. Lett.}\ }\textbf {\bibinfo {volume} {79}},\ \bibinfo {pages}
  {4701} (\bibinfo {year} {1997})}\BibitemShut {NoStop}%
\bibitem [{\citenamefont {Shiino}\ and\ \citenamefont
  {Yoshida}(2001)}]{Shiino2001}%
  \BibitemOpen
  \bibfield  {author} {\bibinfo {author} {\bibfnamefont {M.}~\bibnamefont
  {Shiino}}\ and\ \bibinfo {author} {\bibfnamefont {K.}~\bibnamefont
  {Yoshida}},\ }\href {\doibase 10.1103/PhysRevE.63.026210} {\bibfield
  {journal} {\bibinfo  {journal} {Phys. Rev. E}\ }\textbf {\bibinfo {volume}
  {63}},\ \bibinfo {pages} {026210} (\bibinfo {year} {2001})}\BibitemShut
  {NoStop}%
\bibitem [{\citenamefont {Ichiki}, \citenamefont {Ito},\ and\ \citenamefont
  {Shiino}(2007)}]{Ichiki2007}%
  \BibitemOpen
  \bibfield  {author} {\bibinfo {author} {\bibfnamefont {A.}~\bibnamefont
  {Ichiki}}, \bibinfo {author} {\bibfnamefont {H.}~\bibnamefont {Ito}}, \ and\
  \bibinfo {author} {\bibfnamefont {M.}~\bibnamefont {Shiino}},\ }\href
  {\doibase https://doi.org/10.1016/j.physe.2007.06.042} {\bibfield  {journal}
  {\bibinfo  {journal} {Physica E: Low-dimensional Systems and Nanostructures}\
  }\textbf {\bibinfo {volume} {40}},\ \bibinfo {pages} {402} (\bibinfo {year}
  {2007})}\BibitemShut {NoStop}%
\bibitem [{\citenamefont {Nakao}, \citenamefont {Arai},\ and\ \citenamefont
  {Kawamura}(2007)}]{Nakao2007}%
  \BibitemOpen
  \bibfield  {author} {\bibinfo {author} {\bibfnamefont {H.}~\bibnamefont
  {Nakao}}, \bibinfo {author} {\bibfnamefont {K.}~\bibnamefont {Arai}}, \ and\
  \bibinfo {author} {\bibfnamefont {Y.}~\bibnamefont {Kawamura}},\ }\href
  {\doibase 10.1103/PhysRevLett.98.184101} {\bibfield  {journal} {\bibinfo
  {journal} {Phys. Rev. Lett.}\ }\textbf {\bibinfo {volume} {98}},\ \bibinfo
  {pages} {184101} (\bibinfo {year} {2007})}\BibitemShut {NoStop}%
\bibitem [{\citenamefont {Semenova}\ \emph {et~al.}(2019)\citenamefont
  {Semenova}, \citenamefont {Porte}, \citenamefont {Andreoli}, \citenamefont
  {Jacquot}, \citenamefont {Larger},\ and\ \citenamefont
  {Brunner}}]{Semenova2019}%
  \BibitemOpen
  \bibfield  {author} {\bibinfo {author} {\bibfnamefont {N.}~\bibnamefont
  {Semenova}}, \bibinfo {author} {\bibfnamefont {X.}~\bibnamefont {Porte}},
  \bibinfo {author} {\bibfnamefont {L.}~\bibnamefont {Andreoli}}, \bibinfo
  {author} {\bibfnamefont {M.}~\bibnamefont {Jacquot}}, \bibinfo {author}
  {\bibfnamefont {L.}~\bibnamefont {Larger}}, \ and\ \bibinfo {author}
  {\bibfnamefont {D.}~\bibnamefont {Brunner}},\ }\href {\doibase
  10.1063/1.5120824} {\bibfield  {journal} {\bibinfo  {journal} {Chaos: An
  Interdisciplinary Journal of Nonlinear Science}\ }\textbf {\bibinfo {volume}
  {29}},\ \bibinfo {pages} {103128} (\bibinfo {year} {2019})},\ \Eprint
  {http://arxiv.org/abs/https://doi.org/10.1063/1.5120824}
  {https://doi.org/10.1063/1.5120824} \BibitemShut {NoStop}%
\bibitem [{\citenamefont {Andreoli}\ \emph {et~al.}(2020)\citenamefont
  {Andreoli}, \citenamefont {Porte}, \citenamefont {Chr{\'{e}}tien},
  \citenamefont {Jacquot}, \citenamefont {Larger},\ and\ \citenamefont
  {Brunner}}]{Andreoli2020}%
  \BibitemOpen
  \bibfield  {author} {\bibinfo {author} {\bibfnamefont {L.}~\bibnamefont
  {Andreoli}}, \bibinfo {author} {\bibfnamefont {X.}~\bibnamefont {Porte}},
  \bibinfo {author} {\bibfnamefont {S.}~\bibnamefont {Chr{\'{e}}tien}},
  \bibinfo {author} {\bibfnamefont {M.}~\bibnamefont {Jacquot}}, \bibinfo
  {author} {\bibfnamefont {L.}~\bibnamefont {Larger}}, \ and\ \bibinfo {author}
  {\bibfnamefont {D.}~\bibnamefont {Brunner}},\ }\href {\doibase
  10.1515/nanoph-2020-0171} {\bibfield  {journal} {\bibinfo  {journal}
  {Nanophotonics}\ }\textbf {\bibinfo {volume} {9}},\ \bibinfo {pages} {4139}
  (\bibinfo {year} {2020})}\BibitemShut {NoStop}%
\bibitem [{\citenamefont {Benjamin}\ and\ \citenamefont
  {Cornell}(2014)}]{Benjamin2014}%
  \BibitemOpen
  \bibfield  {author} {\bibinfo {author} {\bibfnamefont {J.~R.}\ \bibnamefont
  {Benjamin}}\ and\ \bibinfo {author} {\bibfnamefont {C.~A.}\ \bibnamefont
  {Cornell}},\ }\href@noop {} {\emph {\bibinfo {title} {Probability,
  Statistics, and Decision for Civil Engineers}}}\ (\bibinfo  {publisher}
  {Dover Publications},\ \bibinfo {year} {2014})\BibitemShut {NoStop}%
\bibitem [{\citenamefont {Jaeger}\ and\ \citenamefont
  {Haas}(2004)}]{Jaeger2004}%
  \BibitemOpen
  \bibfield  {author} {\bibinfo {author} {\bibfnamefont {H.}~\bibnamefont
  {Jaeger}}\ and\ \bibinfo {author} {\bibfnamefont {H.}~\bibnamefont {Haas}},\
  }\href {\doibase 10.1126/science.1091277} {\bibfield  {journal} {\bibinfo
  {journal} {Science}\ }\textbf {\bibinfo {volume} {304}},\ \bibinfo {pages}
  {78} (\bibinfo {year} {2004})},\ \Eprint
  {http://arxiv.org/abs/https://science.sciencemag.org/content/304/5667/78.full.pdf}
  {https://science.sciencemag.org/content/304/5667/78.full.pdf} \BibitemShut
  {NoStop}%
\bibitem [{\citenamefont {Chollet}\ \emph {et~al.}(2015)\citenamefont {Chollet}
  \emph {et~al.}}]{Keras}%
  \BibitemOpen
  \bibfield  {author} {\bibinfo {author} {\bibfnamefont {F.}~\bibnamefont
  {Chollet}} \emph {et~al.},\ }\href@noop {} {\bibfield  {journal} {\bibinfo
  {journal} {GitHub}\ } (\bibinfo {year} {2015})},\ \Eprint
  {http://arxiv.org/abs/https://github.com/fchollet/keras}
  {https://github.com/fchollet/keras} \BibitemShut {NoStop}%
\bibitem [{Bra(2021)}]{Brainjs}%
  \BibitemOpen
  \href@noop {} {\enquote {\bibinfo {title} {brain.js},}\ }\bibinfo
  {howpublished} {\url{https://github.com/BrainJS}} (\bibinfo {year}
  {2021})\BibitemShut {NoStop}%
\bibitem [{\citenamefont {LeCun}(2021)}]{LeCunSite}%
  \BibitemOpen
  \bibfield  {author} {\bibinfo {author} {\bibfnamefont {Y.}~\bibnamefont
  {LeCun}},\ }\href@noop {} {}\bibinfo {howpublished}
  {\url{http://yann.lecun.com/exdb/mnist/index.html}} (\bibinfo {year}
  {2021})\BibitemShut {NoStop}%
\bibitem [{\citenamefont {Mackey}\ and\ \citenamefont
  {Glass}(1977)}]{MackeyGlass1997}%
  \BibitemOpen
  \bibfield  {author} {\bibinfo {author} {\bibfnamefont {M.}~\bibnamefont
  {Mackey}}\ and\ \bibinfo {author} {\bibfnamefont {L.}~\bibnamefont {Glass}},\
  }\href {\doibase 10.1126/science.267326} {\bibfield  {journal} {\bibinfo
  {journal} {Science}\ }\textbf {\bibinfo {volume} {197}},\ \bibinfo {pages}
  {287} (\bibinfo {year} {1977})},\ \Eprint
  {http://arxiv.org/abs/https://science.sciencemag.org/content/197/4300/287.full.pdf}
  {https://science.sciencemag.org/content/197/4300/287.full.pdf} \BibitemShut
  {NoStop}%
\bibitem [{\citenamefont {Freiberger}, \citenamefont {Bienstman},\ and\
  \citenamefont {Dambre}(2020)}]{Freiberger2020}%
  \BibitemOpen
  \bibfield  {author} {\bibinfo {author} {\bibfnamefont {M.}~\bibnamefont
  {Freiberger}}, \bibinfo {author} {\bibfnamefont {P.}~\bibnamefont
  {Bienstman}}, \ and\ \bibinfo {author} {\bibfnamefont {J.}~\bibnamefont
  {Dambre}},\ }\href {\doibase 10.1038/s41598-020-71549-y} {\bibfield
  {journal} {\bibinfo  {journal} {Scientific Reports}\ }\textbf {\bibinfo
  {volume} {10}},\ \bibinfo {pages} {14451} (\bibinfo {year}
  {2020})}\BibitemShut {NoStop}%
\bibitem [{\citenamefont {Murray}(1991)}]{Murray1991}%
  \BibitemOpen
  \bibfield  {author} {\bibinfo {author} {\bibfnamefont {A.}~\bibnamefont
  {Murray}},\ }\href {\doibase 10.1049/el:19910970} {\bibfield  {journal}
  {\bibinfo  {journal} {Electronics Letters}\ }\textbf {\bibinfo {volume}
  {27}},\ \bibinfo {pages} {1546} (\bibinfo {year} {1991})}\BibitemShut
  {NoStop}%
\bibitem [{\citenamefont {Neckar}\ \emph {et~al.}(2019)\citenamefont {Neckar},
  \citenamefont {Fok}, \citenamefont {Benjamin}, \citenamefont {Stewart},
  \citenamefont {Oza}, \citenamefont {Voelker}, \citenamefont {Eliasmith},
  \citenamefont {Manohar},\ and\ \citenamefont {Boahen}}]{Neckar2019}%
  \BibitemOpen
  \bibfield  {author} {\bibinfo {author} {\bibfnamefont {A.}~\bibnamefont
  {Neckar}}, \bibinfo {author} {\bibfnamefont {S.}~\bibnamefont {Fok}},
  \bibinfo {author} {\bibfnamefont {B.~V.}\ \bibnamefont {Benjamin}}, \bibinfo
  {author} {\bibfnamefont {T.~C.}\ \bibnamefont {Stewart}}, \bibinfo {author}
  {\bibfnamefont {N.~N.}\ \bibnamefont {Oza}}, \bibinfo {author} {\bibfnamefont
  {A.~R.}\ \bibnamefont {Voelker}}, \bibinfo {author} {\bibfnamefont
  {C.}~\bibnamefont {Eliasmith}}, \bibinfo {author} {\bibfnamefont
  {R.}~\bibnamefont {Manohar}}, \ and\ \bibinfo {author} {\bibfnamefont
  {K.}~\bibnamefont {Boahen}},\ }\href {\doibase 10.1109/JPROC.2018.2881432}
  {\bibfield  {journal} {\bibinfo  {journal} {Proceedings of the IEEE}\
  }\textbf {\bibinfo {volume} {107}},\ \bibinfo {pages} {144} (\bibinfo {year}
  {2019})}\BibitemShut {NoStop}%
\bibitem [{\citenamefont {Yang}\ \emph
  {et~al.}(2021{\natexlab{a}})\citenamefont {Yang}, \citenamefont {Wang},
  \citenamefont {Zhang}, \citenamefont {Deng}, \citenamefont {Pang},\ and\
  \citenamefont {Azghadi}}]{Yang2021}%
  \BibitemOpen
  \bibfield  {author} {\bibinfo {author} {\bibfnamefont {S.}~\bibnamefont
  {Yang}}, \bibinfo {author} {\bibfnamefont {J.}~\bibnamefont {Wang}}, \bibinfo
  {author} {\bibfnamefont {N.}~\bibnamefont {Zhang}}, \bibinfo {author}
  {\bibfnamefont {B.}~\bibnamefont {Deng}}, \bibinfo {author} {\bibfnamefont
  {Y.}~\bibnamefont {Pang}}, \ and\ \bibinfo {author} {\bibfnamefont {M.~R.}\
  \bibnamefont {Azghadi}},\ }\href {\doibase 10.1109/TNNLS.2021.3057070}
  {\bibfield  {journal} {\bibinfo  {journal} {IEEE Transactions on Neural
  Networks and Learning Systems}\ ,\ \bibinfo {pages} {1}} (\bibinfo {year}
  {2021}{\natexlab{a}})}\BibitemShut {NoStop}%
\bibitem [{\citenamefont {Yang}\ \emph
  {et~al.}(2021{\natexlab{b}})\citenamefont {Yang}, \citenamefont {Gao},
  \citenamefont {Wang}, \citenamefont {Deng}, \citenamefont {Lansdell},\ and\
  \citenamefont {Linares-Barranco}}]{Yang2021b}%
  \BibitemOpen
  \bibfield  {author} {\bibinfo {author} {\bibfnamefont {S.}~\bibnamefont
  {Yang}}, \bibinfo {author} {\bibfnamefont {T.}~\bibnamefont {Gao}}, \bibinfo
  {author} {\bibfnamefont {J.}~\bibnamefont {Wang}}, \bibinfo {author}
  {\bibfnamefont {B.}~\bibnamefont {Deng}}, \bibinfo {author} {\bibfnamefont
  {B.}~\bibnamefont {Lansdell}}, \ and\ \bibinfo {author} {\bibfnamefont
  {B.}~\bibnamefont {Linares-Barranco}},\ }\href {\doibase
  10.3389/fnins.2021.601109} {\bibfield  {journal} {\bibinfo  {journal}
  {Frontiers in Neuroscience}\ }\textbf {\bibinfo {volume} {15}},\ \bibinfo
  {pages} {1} (\bibinfo {year} {2021}{\natexlab{b}})}\BibitemShut {NoStop}%
\bibitem [{\citenamefont {Yang}\ \emph
  {et~al.}(2021{\natexlab{c}})\citenamefont {Yang}, \citenamefont {Wang},
  \citenamefont {Deng}, \citenamefont {Azghadi},\ and\ \citenamefont
  {Linares-Barranco}}]{Yang2021c}%
  \BibitemOpen
  \bibfield  {author} {\bibinfo {author} {\bibfnamefont {S.}~\bibnamefont
  {Yang}}, \bibinfo {author} {\bibfnamefont {J.}~\bibnamefont {Wang}}, \bibinfo
  {author} {\bibfnamefont {B.}~\bibnamefont {Deng}}, \bibinfo {author}
  {\bibfnamefont {M.~R.}\ \bibnamefont {Azghadi}}, \ and\ \bibinfo {author}
  {\bibfnamefont {B.}~\bibnamefont {Linares-Barranco}},\ }\href {\doibase
  10.1109/TNNLS.2021.3084250} {\bibfield  {journal} {\bibinfo  {journal} {IEEE
  Transactions on Neural Networks and Learning Systems}\ ,\ \bibinfo {pages}
  {1}} (\bibinfo {year} {2021}{\natexlab{c}})}\BibitemShut {NoStop}%
\end{thebibliography}%

\end{document}